\documentclass[10pt,twocolumn,letterpaper]{article}

\usepackage{cvpr}
\usepackage{times}
\usepackage{epsfig}
\usepackage{graphicx}
\usepackage{amsmath}
\usepackage{mathrsfs}
\usepackage{amssymb}
\usepackage{url}
\usepackage{caption}
\usepackage{enumitem}
\usepackage{mathrsfs}
\usepackage{algorithm}
\usepackage{algpseudocode}
\usepackage{dsfont}
\usepackage[pagebackref=false,breaklinks=true,letterpaper=true,colorlinks,urlcolor=magenta,citecolor=blue,linkcolor=blue,bookmarks=false]{hyperref}
\usepackage{breakurl}
\usepackage{multirow}
\usepackage{booktabs}
\usepackage{nopageno}

\cvprfinalcopy % *** Uncomment this line for the final submission

\def\cvprPaperID{1280} % *** Enter the CVPR Paper ID here

\newcommand{\figdir}{figures}

\ifcvprfinal\fi
\begin{document}

%%%%%%%%% TITLE
\title{VITAL: VIsual Tracking via Adversarial Learning}

\author{Yibing Song$^1$,~Chao Ma$^2$\thanks{C. Ma is the corresponding author.},~~Xiaohe Wu$^3$,~Lijun Gong$^4$,~Linchao Bao$^1$,\\
Wangmeng Zuo$^3$,~Chunhua Shen$^2$,~Rynson W.H. Lau$^5$,~and Ming-Hsuan Yang$^6$\\\\
$^1$Tencent AI Lab~~~~~$^2$The University of Adelaide~~~~~$^3$Harbin Institute of Technology~~~~~$^4$Tencent\\
$^5$City University of Hong Kong~~~~~$^6$University of California, Merced\\
\small{\url{https://ybsong00.github.io/cvpr18_tracking/index}}
}

\maketitle
%\thispagestyle{empty}

%%%%%%%%% ABSTRACT
\begin{abstract}
The tracking-by-detection framework consists of two stages, i.e., drawing samples around the target object in the first stage and classifying each sample as the target object or as background in the second stage. The performance of existing trackers using deep classification networks is limited by two aspects. First, the positive samples in each frame are highly spatially overlapped, and they fail to capture rich appearance variations. Second, there exists extreme class imbalance between positive and negative samples. This paper presents the VITAL algorithm to address these two problems via adversarial learning. To augment positive samples, we use a generative network to randomly generate masks, which are applied to adaptively dropout input features to capture a variety of appearance changes. With the use of adversarial learning, our network identifies the mask that maintains the most robust features of the target objects over a long temporal span. In addition, to handle the issue of class imbalance, we propose a high-order cost sensitive loss to decrease the effect of easy negative samples to facilitate training the classification network. Extensive experiments on benchmark datasets demonstrate that the proposed tracker performs favorably against state-of-the-art approaches.
\end{abstract}

%%%%%%%%% BODY TEXT
\section{Introduction}
There has been an increasing need for tracking target objects in bounding boxes to understand video contents. Current state-of-the-art trackers are typically based on a two-stage tracking-by-detection framework. The first stage draws a sparse set of samples around the target object and the second stage classifies each sample as either the target object or as the background using a deep neural network. Despite the favorable performance on recent tracking benchmarks~\cite{wu-cvpr13-otb,wu-pami15-otb,Kristan-pami16-vot}, the performance of the two-stage methods is limited by two aspects. First, the positive samples are spatially overlapped, and they cannot capture a variety of appearance changes over time. Second, the extreme foreground-background class imbalance negatively affects training the classification networks. It is of great importance to investigate how to eliminate these barriers to advance the tracking-by-detection framework in the deep learning era.

\renewcommand{\tabcolsep}{.8pt}
\def\swtwo{0.48\linewidth}
\def\swone{0.9\linewidth}
\begin{figure}[t]
\begin{center}
\begin{tabular}{cc}
\vspace{-0.8mm}\includegraphics[width=\swtwo]{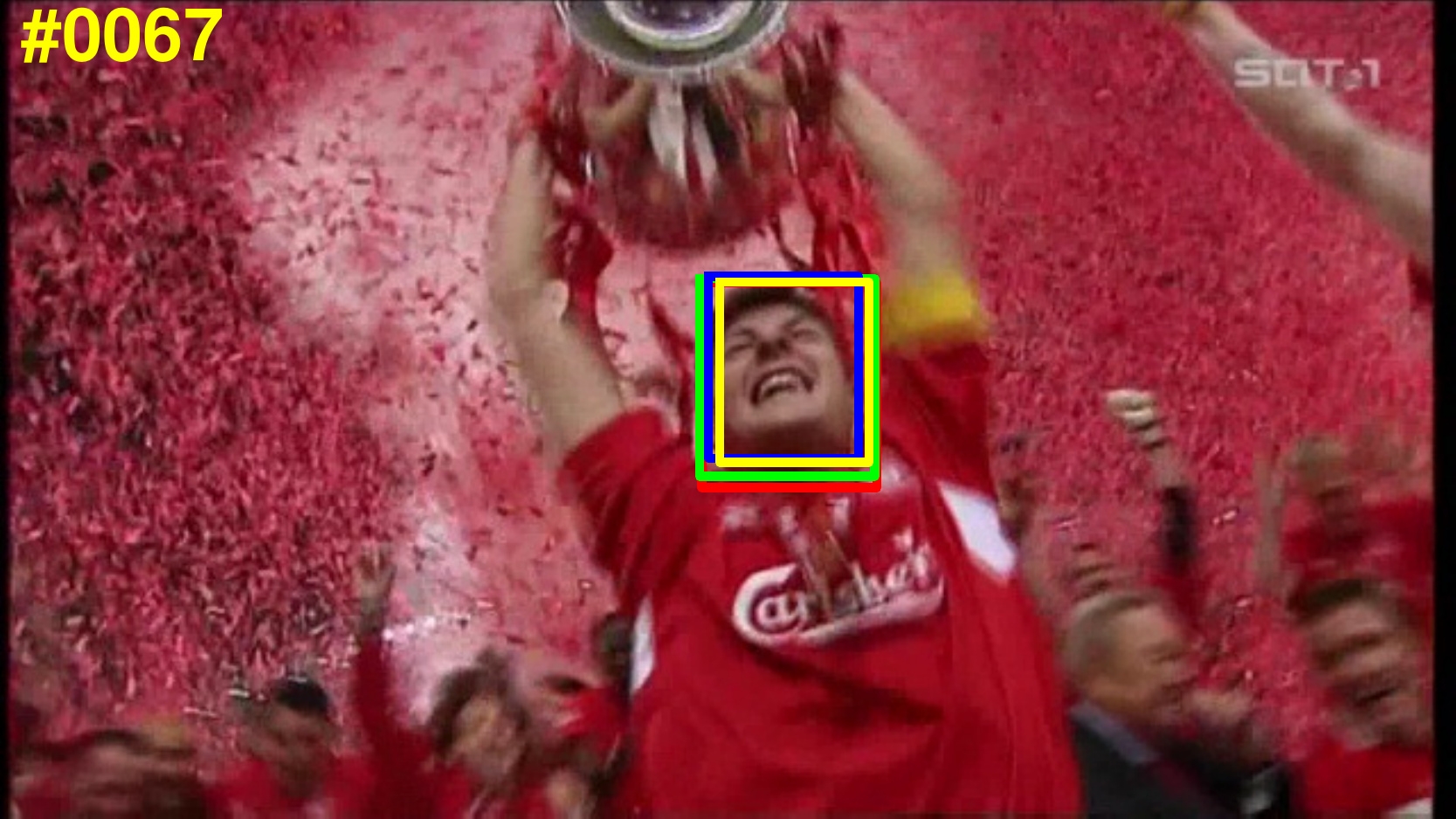}&
\includegraphics[width=\swtwo]{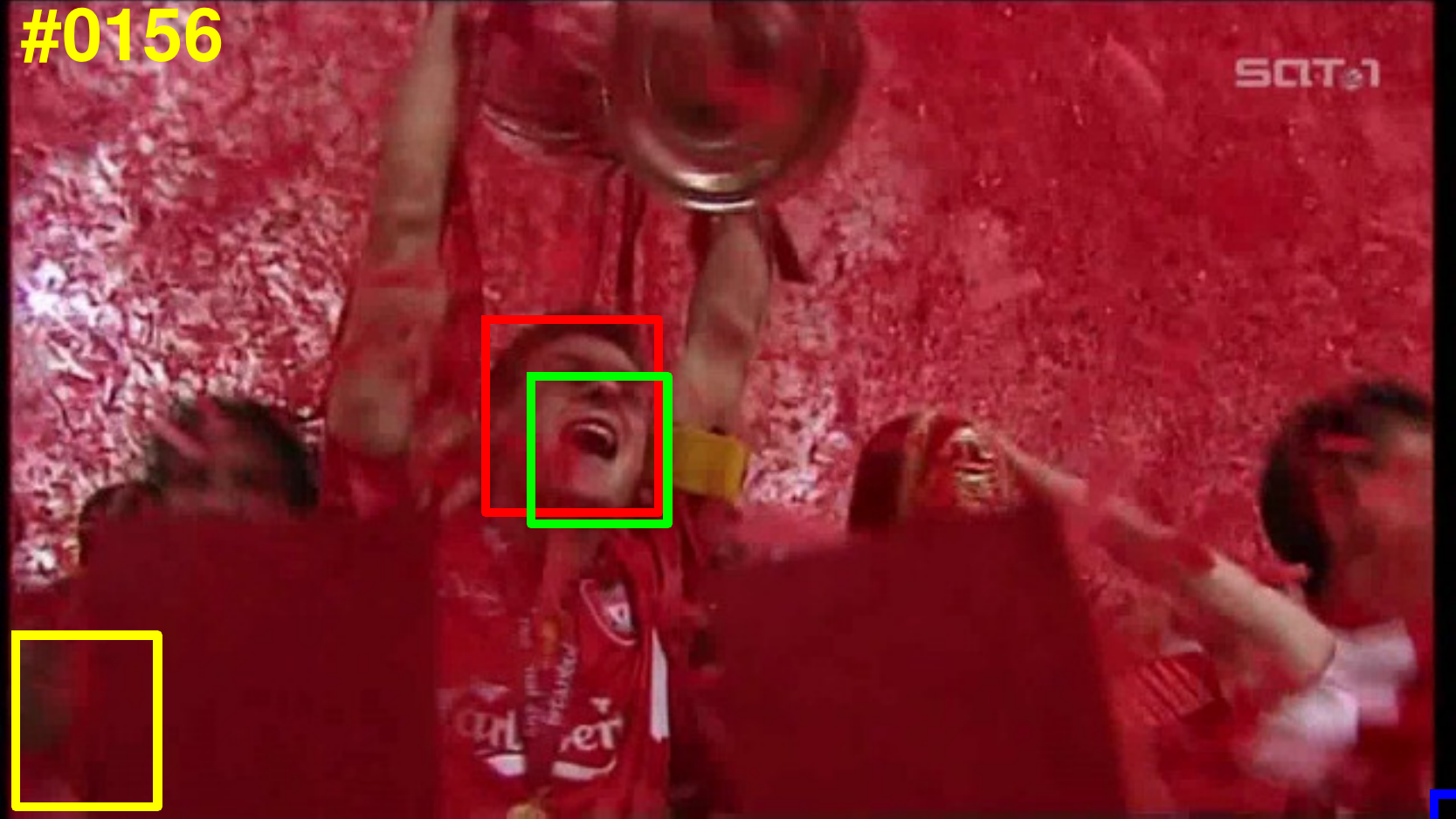}\\
\vspace{-0.8mm}\includegraphics[width=\swtwo]{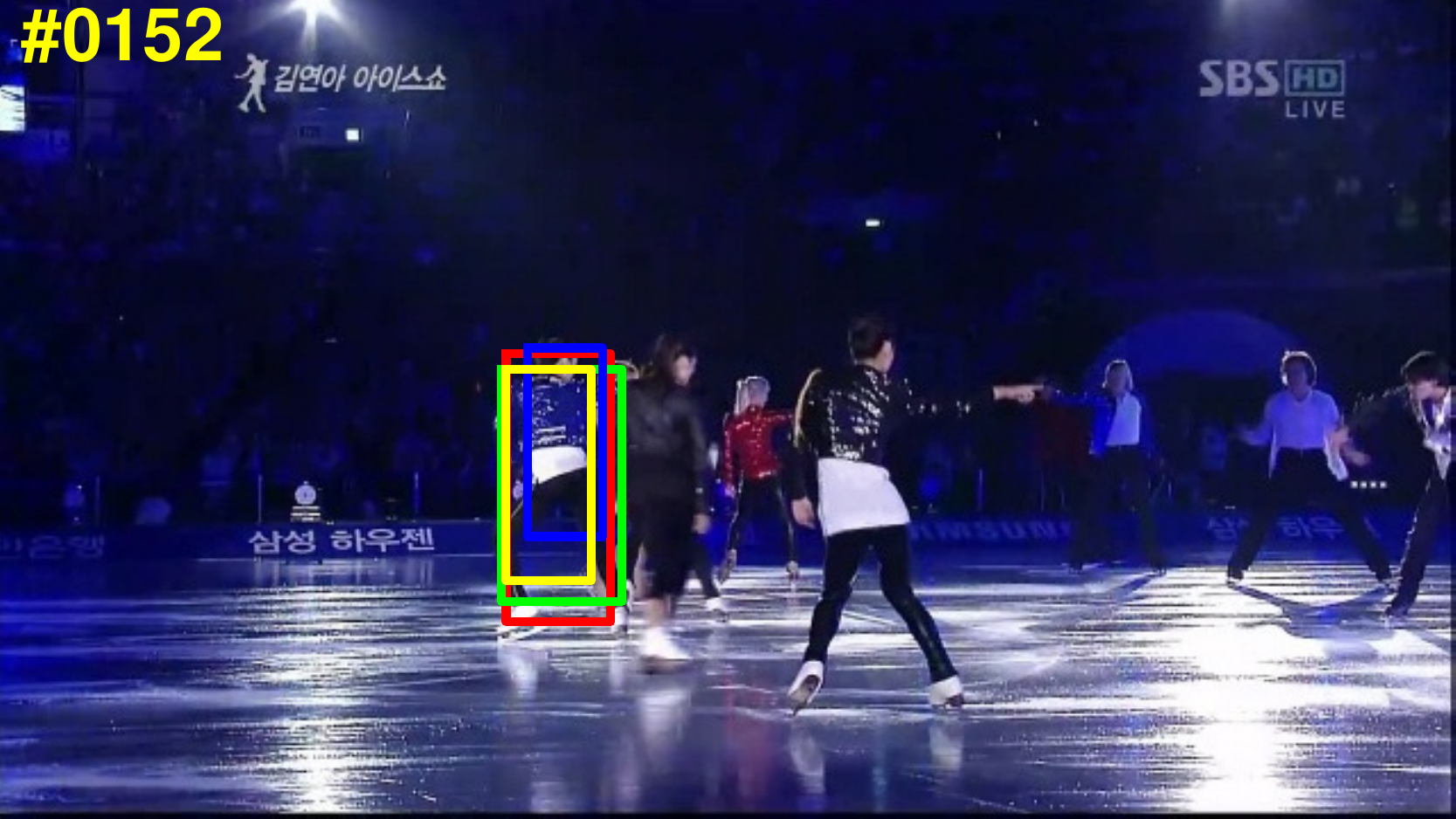}&
\includegraphics[width=\swtwo]{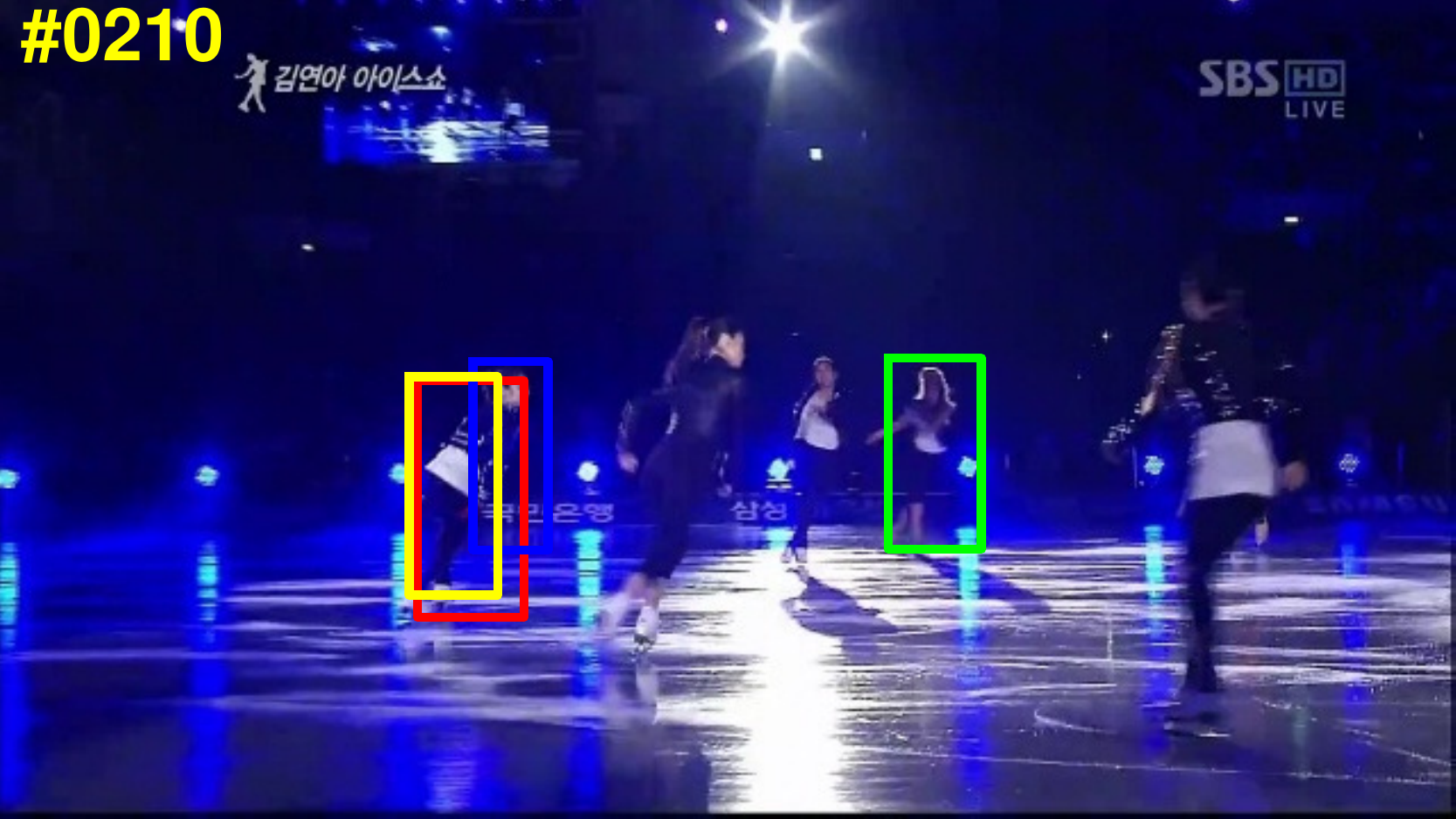}\\
\vspace{-1mm}\includegraphics[width=\swtwo]{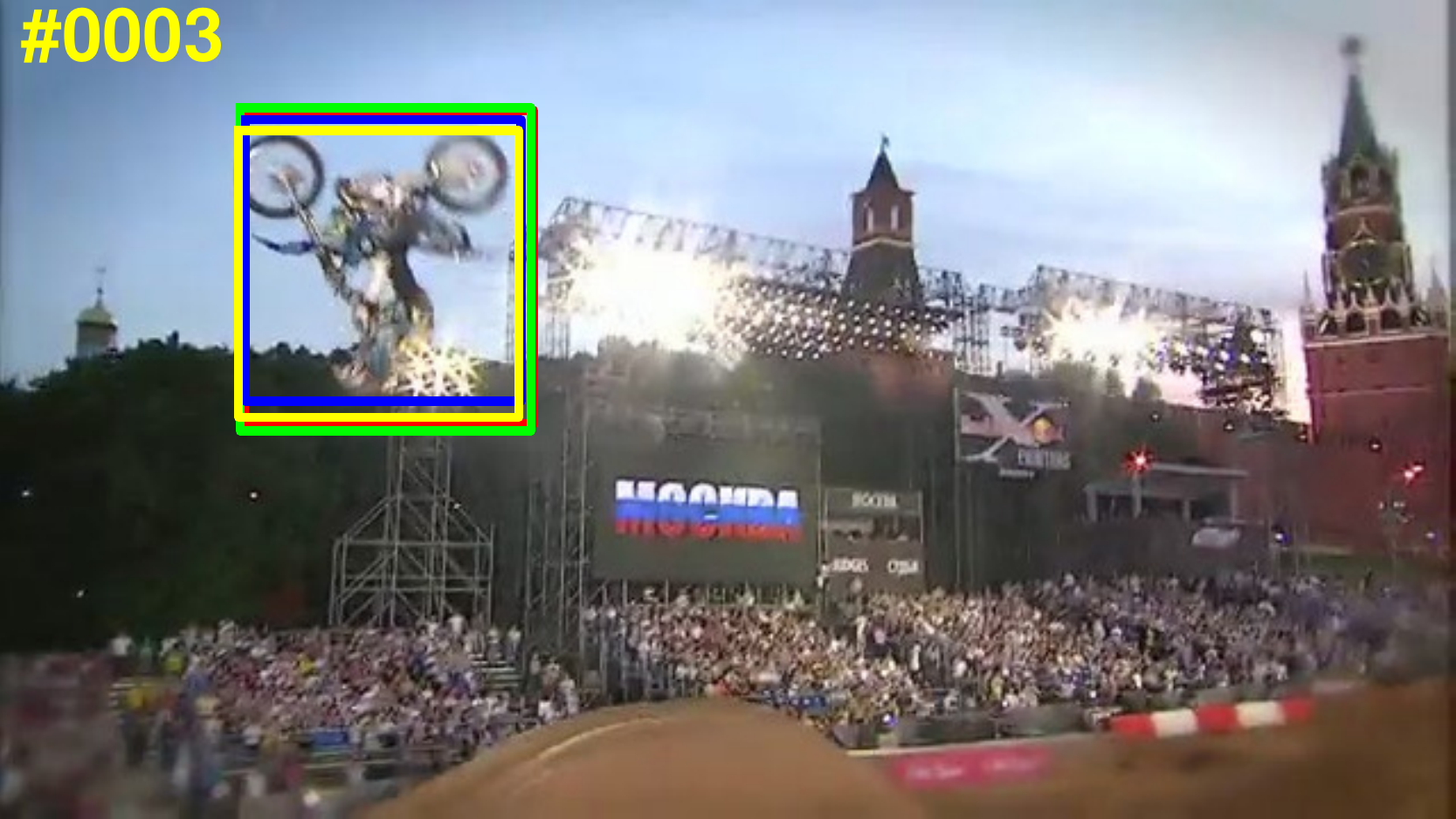}&
\includegraphics[width=\swtwo]{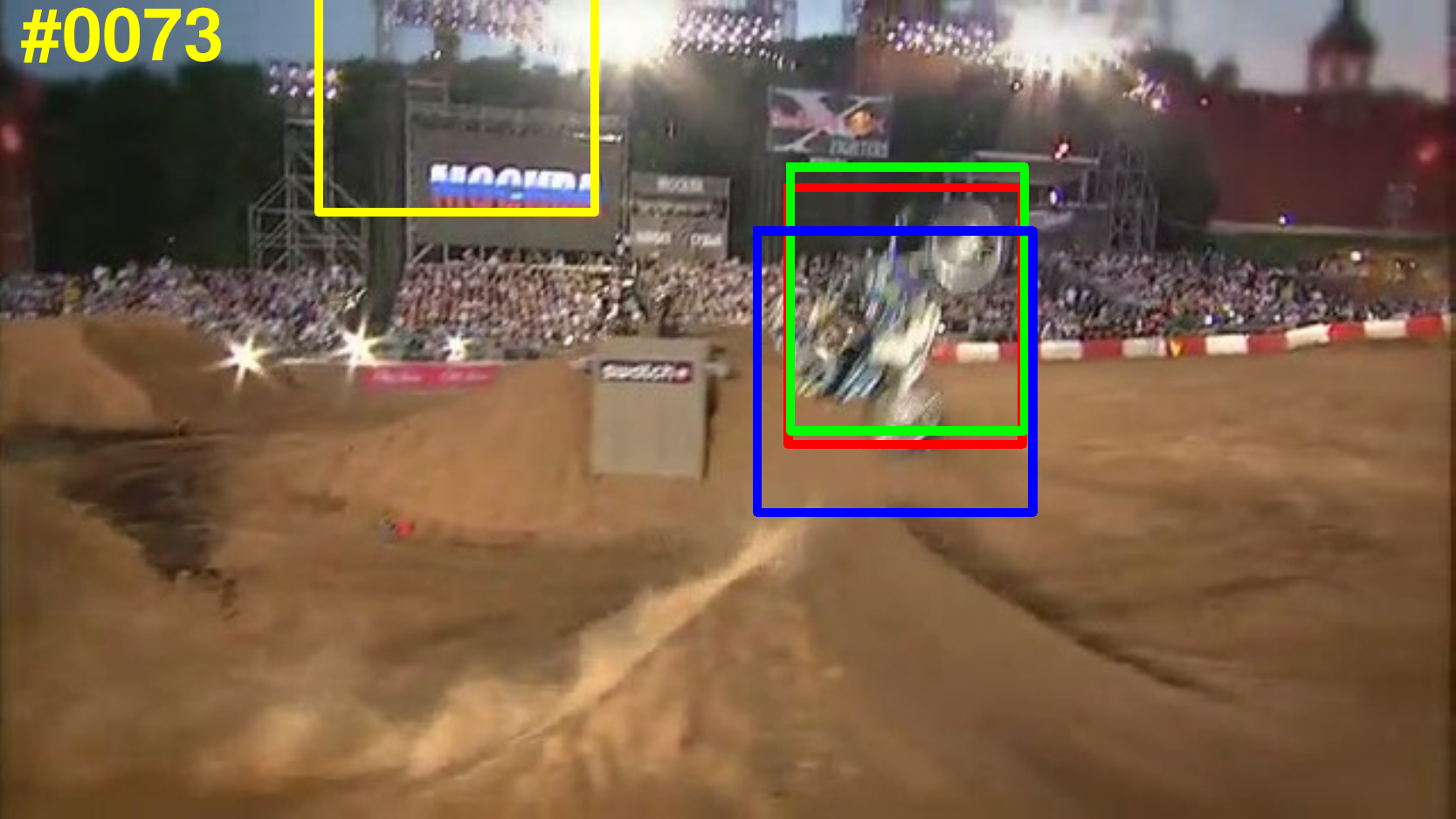}\\
\end{tabular}
\begin{tabular}{c}
\includegraphics[width=\swone]{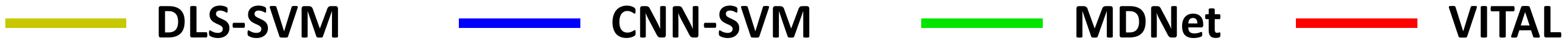}\\
\end{tabular}
\end{center}
\vspace{-5mm}
\caption{Tracking results with the comparison to state-of-the-art tracking-by-detection trackers including DLS-SVM~\cite{ning-cvpr16-object}, CNN-SVM~\cite{hong-icml15-cnnsvm}, and MDNet~\cite{nam-cvpr16-mdnet}. Our VITAL tracker learns to diversify positive samples via adversarial learning and to balance training samples via cost sensitive loss. It performs favorably against existing trackers.}
\label{fig:intro}
\end{figure}

Prior trackers have made limited efforts on increasing the diversity of training data in learning deep classifiers. Since classifiers tend to learn a discriminative boundary between positive and negative samples, they emphasize on the most discriminative ones. However, as the target appearance varies frame-by-frame in the whole video sequence, the most discriminative samples in the current frame may not persist over a long temporal span. Typical examples of appearance changes caused by partial occlusion or out-of-plane rotation easily result in model overfitting, as current training samples may differ much from the previous ones. To alleviate this problem, existing trackers incrementally update the classifier through online sample collections. The noisy updates occur and bring tracker drift problem. Hence, a natural question is how we can augment positive samples in the feature space to capture target appearance variations in the temporal domain.

In this work, we take advantage of the recent progress in adversarial learning to augment training data to facilitate classifier training. For a deep classification network, such as the VGG-M model~\cite{simonyan-iclr14-very}, we add a generative network between the last convolutional layer and the first fully connected layer. The generative network augments positive samples by generating weight masks randomly applied to the features, where each mask represents a specific type of appearance variation. Through adversarial learning, our network can identify the mask that maintains the most robust features of target appearance in the temporal domain. We show that the learned mask tends to decrease the weights of discriminative features, which tends to overfit in a single frame. Meanwhile, these features are hardly robust to appearance changes over the temporal span. In other words, adversarial learning helps our tracker exploit the most robust features over a long temporal span in classifier training, rather than overfitting to discriminative features in a single frame. Moreover, to mitigate the issue of class imbalance, we propose a high-order cost sensitive loss to decrease the effect of easy negative samples. Taking advantages of adversarial learning and high-order cost sensitive loss, our tracking method achieves favorable results against state-of-the-art trackers.

We summarize the main contributions of this work as follows:
\begin{itemize}[noitemsep,nolistsep]
  \item We propose to use a generative adversarial network (GAN) to augment positive samples in the feature space to capture a variety of appearance changes over a temporal span.
  \item We propose to use higher-order cost sensitive loss to mine hard negative samples to handle class imbalance.
  \item We extensively validate our method on benchmark datasets with large-scale sequences. We show that our VITAL tracker performs favorably against state-of-the-art trackers.
\end{itemize}

\section{Related Work}
Visual tracking has long been an active research topic with extensive surveys~\cite{smeulders-pami14-visual} over the last decade. In this section, we mainly discuss the representative visual trackers and the related issues on generative adversarial learning and class imbalance.

{\flushleft \bf Visual Tracking.} Visual tracking has a wide range of applications including action recognition~\cite{choi-eccv12-unified}, target analysis~\cite{song-ijcai17-faceSR,song-ijcai17-faceSketch,song-cviu17-style} and augmented reality~\cite{comport-tcvg06-real,ren-eccv16-dehaze}. State-of-the-art trackers are mainly based on the one-stage regression framework or the two-stage classification framework. As one of the most representative types of the one-stage regression framework, the correlation filter based trackers regress all the circular-shifted version of the input features into soft labels generated by a Gaussian function. By computing the correlation as an element-wise product in the Fourier domain, these trackers have received a lot of attention recently. Starting from the MOSSE tracker \cite{bolme-cvpr10-mosse}, many efforts have been made to improve the correlation filter for robust tracking. Extensions include, but are not limited to, kernelized correlation filters \cite{henriques-pami15-high}, scale estimation \cite{martin-bmvc14-accurate}, re-detection \cite{ma-cvpr15-lct}, spatial regularization \cite{martin-iccv15-learning,martin-eccv16-beyond,martin-cvpr17-eco}, ADMM optimization~\cite{kiani-cvpr15-correlation}, sparse representation~\cite{lan-tip15-joint,Qi-TIP18-sparse,lan-tip18-learning}, CNN feature integrations~\cite{chao-iccv15-HCF,qi-cvpr16-hdt,zhang-cvpr17-mcpf,kiani-iccv17-learning} and end-to-end CNN predictions~\cite{wang-iccv15-visual,luca-cvpr17-end,song-iccv17-CREST}.

In contrast, the two-stage classification framework poses the tracking task as a binary classification problem. The two-stage trackers emphasize on a discriminative boundary between the samples of the target object and background. Numerous learning schemes are proposed including P-N learning~\cite{kalal-pami12-tld}, multiple instance learning~\cite{babenko-pami11-mil}, structured SVMs~\cite{hare-pami16-struck,ning-cvpr16-object}, CNN-SVMs~\cite{hong-icml15-cnnsvm}, domain adaptation~\cite{nam-cvpr16-mdnet}, and ensemble learning~\cite{han-cvpr17-branchout}. Unlike the existing two-stage tracking-by-detection trackers, our method, for the first time, takes advantage of the recent progress in generative adversarial learning to augment training samples in the feature space. The augmented samples capture a variety of appearance changes and thus strengthen the robustness of the classifier. In addition, we exploit hard negative samples to handle class imbalance limitation.

\begin{figure*}[t]
	\begin{center}
		\begin{tabular}{c}
			\includegraphics[width=0.9\linewidth]{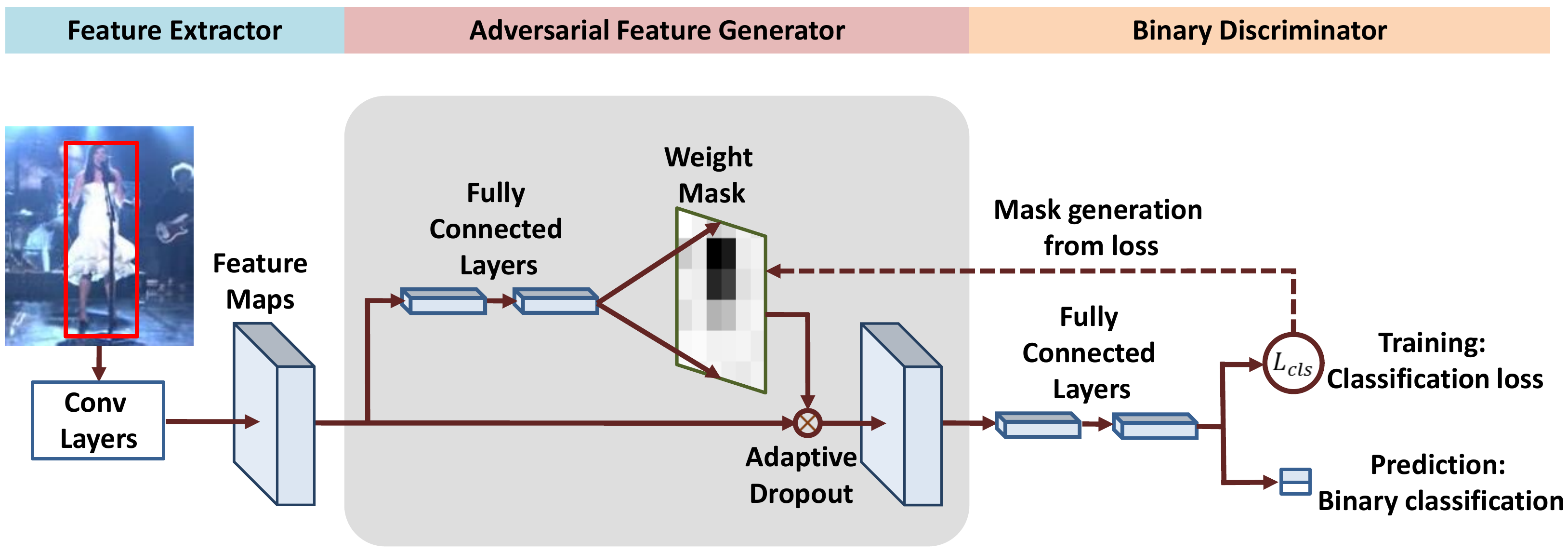}\\
		\end{tabular}
	\end{center}
	\vspace{-5mm}
	\caption{Overview of our network architecture. Our method takes each sampled patch as an input and predicts a possibility score of the patch being the target object. We add one branch of fully connected layers after the last convolutional layer to randomly generate masks, which are applied to the input features to capture a variety of appearance changes. Adversarial learning identifies the mask that maintains the most robust features over a long temporal span while removing the discriminative features from individual frames. It facilitates the classifier training process.}
	\label{fig:pipeline}
\end{figure*}

{\flushleft \bf Generative Adversarial Learning.}  It is introduced in~\cite{goodfellow-nips14-gan} to generate realistic-looking images from random noise via the CNN. The generative adversarial network (GAN) consists of two subnetworks. One serves as a generator and the other as a discriminator. The generator aims at synthesizing images to fool the discriminator, while the discriminator tries to discriminates between real images and images synthesized by the generator. The generator and the discriminator are trained simultaneously by competing with each other. An advantage of adversarial learning is that the generator is trained to produce similar image statistics to those of the training samples so that the discriminator cannot differentiate. This manner is hardly achieved by existing empirical objective functions with supervised learning. The progress in generative adversarial learning has attracted a series of works on network training~\cite{Martin-icml17-wGan,nowozin-nisp16-fGan,gurumurthy-cvpr17-deligan} and computer vision applications, such as image generation~\cite{zhang-iccv17-stackgan}, image stylization~\cite{isola-cvpr17-cGAN}, object detection~\cite{wang-cvpr17-fast}, and semantic segmentation~\cite{souly-iccv17-semi}. Unlike existing GANs that augment data in the image space, we apply adversarial learning to augment training samples in the feature space to capture appearance variations in temporal domain. In sum, our method exploits robust features over the long temporal span, instead of the discriminative features in individual frames.

{\flushleft \bf Class Imbalance.} This problem often exists in learning applications, where the amount of training data in one class (usually the positive class) is far less than that of another class (usually the negative class). A large portion of samples from the majority class are easy samples, which dominantly produce a large loss, and make the learning process unaware of the valuable samples from the minority class. Hard negative mining~\cite{felzenszwalb-cvpr10-cascade,shrivastava-cvpr16-training} and reweighing training data~\cite{rota-cvpr17-loss,lin-iccv17-focal} are useful to alleviate the class imbalance problem to some extent. In visual tracking, class imbalance deteriorates the performance of the classifier, as the number of positive samples are extremely limited but the number of negative samples across the whole background is large. Unlike the aforementioned solutions for the class imbalance problem, we propose cost sensitive loss to decrease the effect from easy negative samples when training the classifier. This not only improves the tracking accuracy, but also accelerates the training convergence.

\section{Proposed Algorithm}
We build VITAL upon the CNN tracking-by-detection framework, which consists of feature extraction and classification. We interpret the classifier as the discriminator and propose a generator for adversarial learning~\cite{goodfellow-nips14-gan}. Unlike existing GAN-based methods, which expect to obtain generator mapping samples from one distribution to another after the training process, we expect to obtain a discriminator which is robust to target object variations. Fig. \ref{fig:pipeline} shows the pipeline of our method, and the details are discussed below.

\subsection{Adversarial Learning}\label{sec:al}
In the traditional adversarial learning~\cite{goodfellow-nips14-gan}, the generator $G$ takes a noise vector $z$ from a distribution $P_{noise}(z)$ as an input and outputs an image $G(z)$. The discriminator $D$ takes either $G(z)$ or a real image $x$ with a distribution $P_{data}(x)$ as an input and outputs the classification probability. The generator $G$ is learned to maximize the probability of $D$ making a mistake. Using the standard cross entropy loss, the objective loss function for training $G$ and $D$ is defined as:
\begin{equation}
\begin{aligned}
\mathcal{L}=\min \limits_{G} \max \limits_{D}~~&\mathbb{E}_{x\backsim P_{data}(x)}[\log D(x)]\\
+~&\mathbb{E}_{z\backsim P_{noise}(z)}[\log (1-D(G(z)))],
\end{aligned}
\label{eq:gan}
\end{equation}
where the $G$ and $D$ networks are trained simultaneously. The training encourages $G$ to fit $P_{data}(x)$ so that $D$ will not be able to discriminate $x$ from $G(z)$. Note that in Eq.~\ref{eq:gan}, there are no ground truth annotations for $z$ and the learning process is unsupervised. After the training process, $G$ is removed and only $D$ is kept for inference.

Although GANs have been investigated in many computer vision tasks, a direct applying of Eq.~\ref{eq:gan}
in the tracking-by-detection framework is not feasible. First, the input data to the framework are usually candidate object proposals rather than random noise. Second, we need to train the classifier via supervised learning using labeled samples rather than unlabeled ones. Third, we expect to use the classifier (i.e., $D$) for inference rather than $G$. These three factors limit the usage of GANs on visual tracking where both the input and learning strategy differ significantly.

We propose VITAL to narrow the gap between GANs and the tracking-by-detection framework. We add $G$ between feature extraction and the classifier as shown in Fig. \ref{fig:pipeline}. $G$ will predict a weight mask which operates on the extracted features. This mask is set randomly at the beginning and gradually identifies the discriminative features through adversarial learning. We define the input feature as $C$, the mask generated by the $G$ network as $G(C)$, the actual mask identifying the discriminative features as $M$. We define the objective function as:
\begin{equation}
\begin{aligned}
\mathcal{L}_{\mathrm{VITAL}}=\min \limits_{G} \max \limits_{D}~&\mathbb{E}_{(C,M)\backsim P_{(C,M)}}[\log D(M\cdot C)]\\
+&\mathbb{E}_{C\backsim P_{(C)}}[\log (1-D(G(C)\cdot C))]\\
+&\lambda \mathbb{E}_{(C,M)\backsim P_{(C,M)}}||G(C)-M||^2,
\end{aligned}
\label{eq:ganVital}
\end{equation}
where the dot is the dropout operation on the feature $C$. The mask contains only one channel and has the same resolution as $C$. We express the predicted mask as $\hat{M}$ and the value of the element $(i,j)$ as $\hat{M_{ij}}$. Meanwhile, we define the value of the element $(i,j,k)$ on feature $C$ as $C_{ijk}$. The dropout operation is defined as follows:
\begin{equation}
C^{\mathrm{o}}_{ijk}=C_{ijk} \hat{M_{ij}},
\label{eq:maskout}
\end{equation}
where $C^{\mathrm{o}}_{ijk}$ is the feature $C$ after the dropout operation and passed onto the classifier.

In Eq.~\ref{eq:ganVital}, we integrate the adversarial learning into the tracking-by-detection framework.
We keep the input (i.e., the candidate object proposals) unchanged. When training $D$ (i.e, classifier), we extract features and enrich their representations in the feature space. Instead of empirically proposing data augmentation strategies, we let $G$ to identify the discriminative features, which are crucial for training $D$. Initially, $G$ produces several random masks, which are akin to the random noise in Eq.~\ref{eq:gan}. Each mask represents a specific type of appearance variation, and we expect these masks to cover the whole object variations. Through the adversarial learning process, $G$ will gradually identify the mask that degrades the classifier most. This indicates that the mask has identified the discriminative features. On the other hand, $D$ will gradually be trained without overfitting to the discriminative features from individual frames while relying on more robust features over a long temporal span. In each iteration of the adversarial learning, we first train $D$ and then $G$. The detailed training procedure is presented in the following:

{\flushleft \bf Training $D$.} In one iteration of the training process, we pass the input feature through $G$ and obtain the predicted mask $\hat{M}$. We then conduct the dropout operation on this feature and sent the modified feature into $D$. We keep the labels unchanged and train $D$ through supervised learning. Note that during this training process, there are multiple input features, $G$ will predict different masks according to different input features. It enables $D$ to focus on the temporal robust features without discriminative feature interference from single frames.

{\flushleft \bf Training $G$.} After training $D$ once, given an input feature, we create multiple output features based on several random masks. This feature diversifying process is performed through the dropout operation illustrated in Eq. \ref{eq:maskout}. These features are passed onto $D$, and we pick up the one with the highest loss. The corresponding mask of the selected feature is said to be effective in decreasing the impact of the discriminative features. We set this mask as $M$ in Eq.~\ref{eq:ganVital} and update $G$ accordingly.

\renewcommand{\tabcolsep}{.8pt}
\def\swthree{0.32\linewidth}
\def\swone{0.8\linewidth}
\begin{figure}[t]
\small	
\begin{center}
\begin{tabular}{c}
	%\vspace{-2mm}
	Entropy intensity bar \\
	\includegraphics[width=\swone]{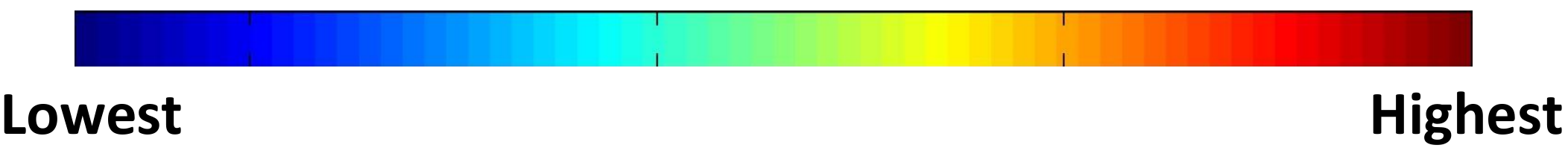}\\
\end{tabular}
\begin{tabular}{ccc}
\vspace{-1mm}\includegraphics[width=\swthree]{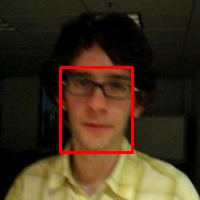}&
\includegraphics[width=\swthree]{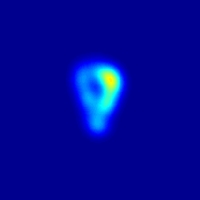}&
\includegraphics[width=\swthree]{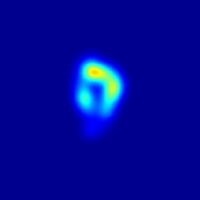}\\
(a) Input &(b) Entropy map&(c) Entropy map\\
Frame \#303 & without GAN & with GAN\\
\vspace{-1mm}\includegraphics[width=\swthree]{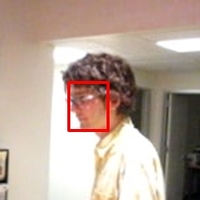}&
\includegraphics[width=\swthree]{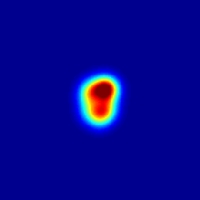}&
\includegraphics[width=\swthree]{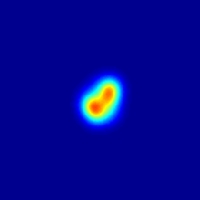}\\
(d) Input &(e) Entropy map&(f) Entropy map\\
Frame \#460 & without GAN & with GAN\\
\end{tabular}
\end{center}
\vspace{-5mm}
\caption{Entropy distribution of two frames on the \emph{David} sequence~\cite{wu-cvpr13-otb}. We use VITAL with and without GAN integration for comparison. We analyze the entropy based on the predicted probabilities from the classifier. The higher the entropy, the more uncertain the classifier prediction is.}
\label{fig:algo}
\end{figure}

{\flushleft \bf Visualization.} Adversarial learning enables the classifier to focus on the temporal robust features instead of the discriminative ones in individual frames. Fig. \ref{fig:algo} shows an example of how adversarial learning affects the classifier in practice. Fig. \ref{fig:algo}(a) shows the input frame with the ground truth annotation located at the face region. We use our VITAL tracker to represent the tracking-by-detection framework for illustration. We compute the entropy distribution based on the predicted probabilities from the classifier. The entropy measures the uncertainty of the prediction and is computed for binary classification as:
\begin{equation}
H=-\big(p\cdot \log p+(1-p)\cdot \log(1-p)\big),
\label{eq:entropy}
\end{equation}
where $p$ is the predicted probability of the target object and $1-p$ is the background. When $p=0.5$, the value of the entropy $H$ is highest, which means that the classifier is uncertain to predict the label. When $p=0$ or $p=1$, the value of the entropy $H$ is lowest, which means that the classifier is certain about the prediction.

We compute the entropy distribution of Fig. \ref{fig:algo}(a) using VITAL without adversarial learning as shown in Fig. \ref{fig:algo}(b) and with adversarial learning as shown in Fig. \ref{fig:algo}(c). We note that these two distributions are similar despite some tiny variances. However, when the target undergoes partial occlusion and out-of-plane rotation as shown in Fig. \ref{fig:algo}(d), the entropy of VITAL without adversarial learning increases rapidly as shown in Fig. \ref{fig:algo}(e), which indicates that the classifier becomes uncertain around the target region. This is because the classifier is trained to focus on the discriminative features of the samples in the previous frames. As the target appearance varies in the following frames, these discriminative features vanish and decrease the classification accuracy. In comparison, the entropy distribution shown in Fig. \ref{fig:algo}(f) does not vary as significant as that in Fig. \ref{fig:algo}(e). It is because the classifier trained via diversified samples will not focus on the most discriminative features in individual frames. Instead, it tends to focus on more robust features over a long period of time. In sum, with the adversarial learning, VITAL becomes temporally robust while preserving the classification accuracy on individual frames.

\subsection{Cost Sensitive Loss}

We first revisit the cross entropy (CE) loss for binary classification. Formally, we define $y\in\{0,1\}$ as the class labels and $p\in[0,1]$ as the estimated probability for a class with label $y=1$. Meanwhile, we define the probability for a class with label $y=0$ as $1-p$. The CE loss is formulated as:
\begin{equation}
L(p,y)=-\big(y\cdot\log(p)+(1-y)\cdot\log(1-p)\big).
\end{equation}
One notable problem of the CE loss is that easy negative samples, i.e., when $p\ll0.5$ and $y=0$, produce the loss with non-trivial magnitude. When summed over a large number of easy negative examples, these
small loss values overwhelm the valuable rare positive class. In visual tracking, class imbalance lies between the limited positive samples and a substantial amount of negative samples across the whole background. Easy negative samples take over the majority of the CE loss and dominate the gradient.

Existing solutions to class imbalance include hard negative mining~\cite{felzenszwalb-cvpr10-cascade,shrivastava-cvpr16-training} and training data reweighing~\cite{rota-cvpr17-loss}. The simplest method to make a classifier cost sensitive involves a modification of the class importance. For example, when the ratio of positive and negative classes is 1:100, the importance factor of the negative class is set to be 0.01.
Note that simply using a fixed factor to balance the importance of positive/negative examples does not identify the easiness or hardness of each example. We align our motivation to the recently proposed focal loss~\cite{lin-iccv17-focal} and add a modulating factor to the CE loss in terms of the network output probability $p$. Formally, we build our cost sensitive loss upon the entropy loss as:
\begin{equation}
L(p,y)=-\big(y\cdot (1-p)\cdot\log(p)+(1-y)\cdot p\cdot\log(1-p)\big).
\end{equation}
With the cost sensitive loss, we reformulate the objective function in Eq. \ref{eq:ganVital} as:
\begin{equation}
\begin{aligned}
\mathcal{L}_{\mathrm{VITAL}}=&\min \limits_{G} \max \limits_{D}~\mathbb{E}_{(C,M)\backsim P_{(C,M)}}[K_1\cdot\log D(M\cdot C)]\\
&~~+\mathbb{E}_{C\backsim P_{(C)}}[K_2\cdot\log (1-D(G(C)\cdot C))]\\
&~~+\lambda \mathbb{E}_{(C,M)\backsim P_{(C,M)}}||G(C)-M||^2,
\end{aligned}
\label{eq:ganVitalCSR}
\end{equation}
where $K_1=1-D(M\cdot C)$ and $K_2=D(G(C)\cdot C)$ are modulating factors that balance the training sample loss.

\section{Tracking via VITAL}
We illustrate how we perform VITAL for visual tracking. Note that we only involve $G$ when training the classifier and remove it in the test stage. The details are as follows:

{\flushleft \bf Model initialization.} We initialize our model through a two-stage training. In the first step we offline pretrain the model using positive and negative samples from the training data, which is from \cite{nam-cvpr16-mdnet}. In the second step we draw the samples from the first frame of the input sequence to finetune our model online. During offline pretraining, we randomly initialize $D$ and perform the training in a few iterations, then we involve $G$ for adversarial learning. See Sec. \ref{sec:al} for the details of the adversarial learning process where only positive samples are adopted. We mine the hard negative samples through the cost sensitive loss for training $D$ together with the diversified positive samples.

{\flushleft \bf Online detection.} The online detection scheme is the same as existing tracking-by-detection approaches as we remove $G$ in this step. Given an input frame, we first generate multiple candidate proposals and extract their CNN features. We feed the CNN features of the candidate proposals into the classifier to get the probability scores.

{\flushleft \bf Model update.} We incrementally update our tracker frame-by-frame. Around the estimated position, we generate multiple samples and assign them with binary labels according to their intersection-over-union scores with the estimated bounding box. We use these training samples jointly train $G$ and $D$ during online update as illustrated in Sec. \ref{sec:al}.

\renewcommand{\tabcolsep}{0.1pt}
\begin{figure}[t]
\begin{center}
\begin{tabular}{cc}
\includegraphics[width=\swtwo]{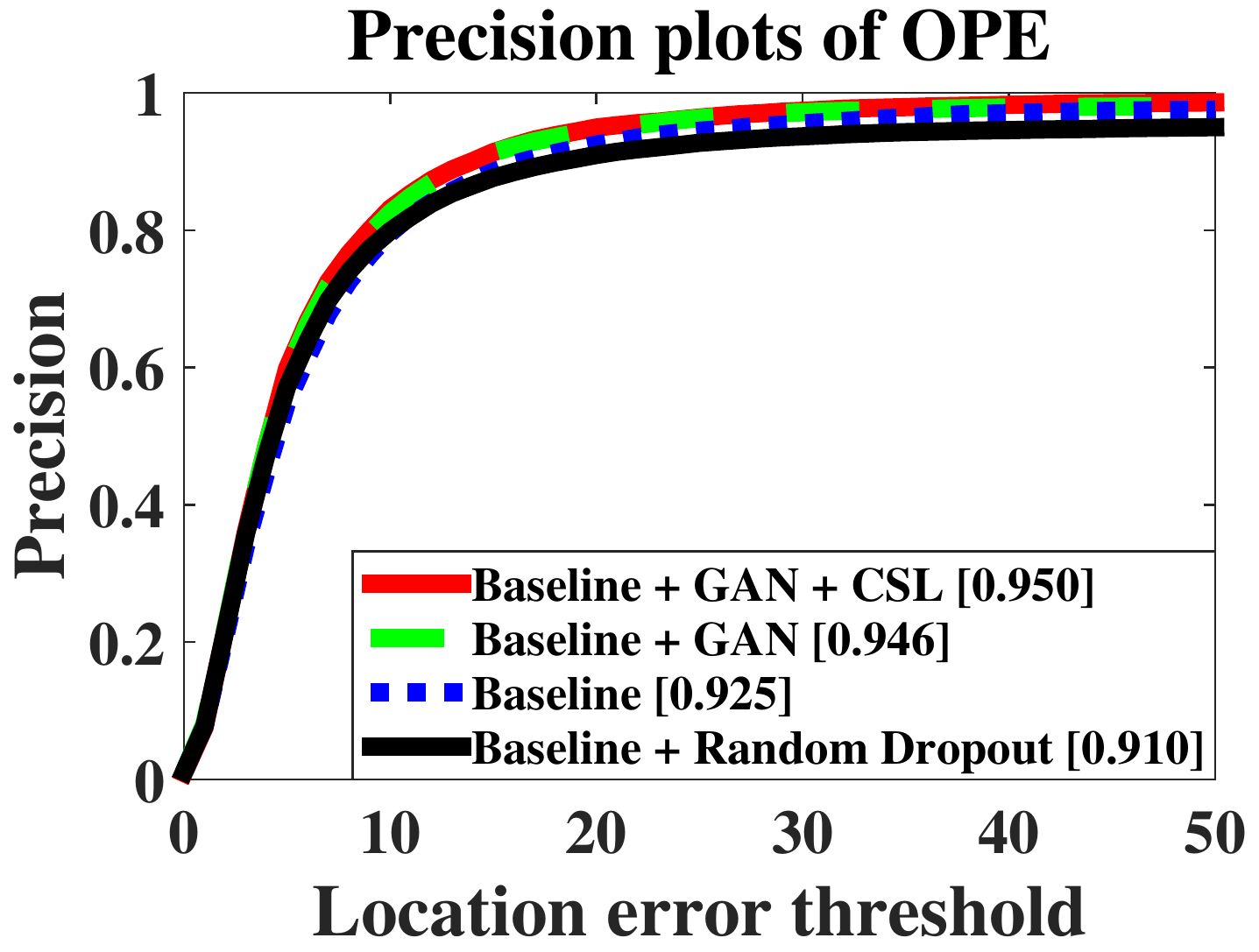}&
\includegraphics[width=\swtwo]{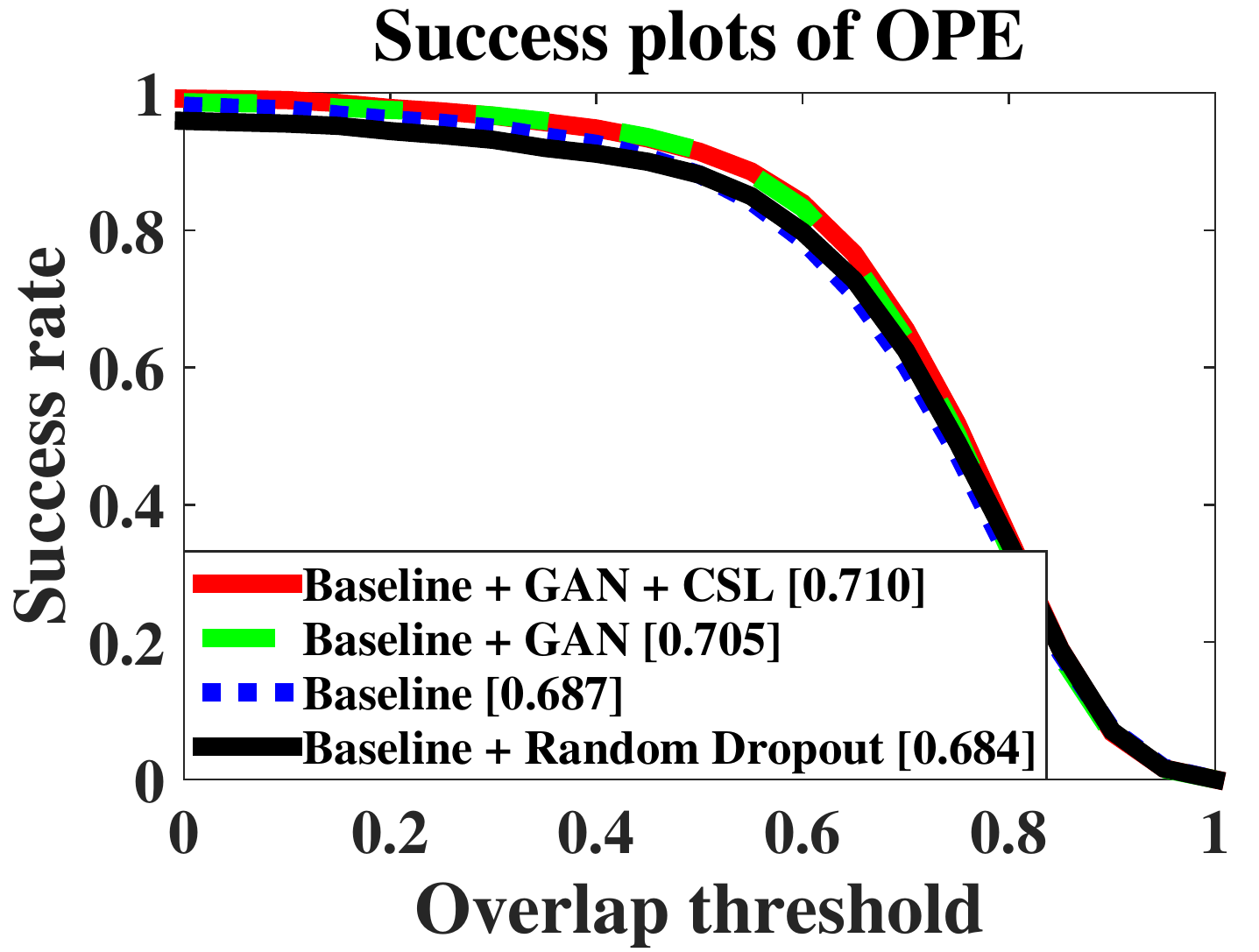}\\
\end{tabular}
\end{center}
\vspace{-5mm}
\caption{Precision and success plots on the OTB-2013 dataset using the one-pass evaluation. The numbers in the legend indicate the average distance precision scores at 20 pixels and the area-under-the-curve success scores.}
\label{fig:ablation}
\end{figure}

\section{Experiments}
In this section, we introduce the implementation details of VITAL and analyze the effects of adversarial learning and cost sensitive loss. Then we compare our VITAL tracker with state-of-the-art trackers on the benchmark datasets OTB-2013~\cite{wu-cvpr13-otb}, OTB-2015~\cite{wu-pami15-otb} and VOT-2016~\cite{kristan-eccvw16-vot} for performance evaluation.

{\flushleft \bf Experimental Setup.}
Our backbone feature extractor is based on the first three convolutional layers from the VGG-M model~\cite{simonyan-iclr14-very}. When training $G$, we prepare 9 random masks. The resolution of each mask is the same as that of the input features. We split this mask into 9 parts equally. We assign each part with label 1 in turn and the remaining parts with label 0. These masks are different from each other and cover all the parts in total. When training $D$, we apply 9 masks to the input features independently to generate 9 diversified versions of each input feature. We then feed these diversified features into $D$ and select the one with the highest loss. The corresponding mask is denoted by $M$ as illustrated in Eq. \ref{eq:ganVital} to train $D$. During the adversarial learning, we iteratively apply the SGD solver to both $G$ and $D$. We use 100 iterations to initialize both networks. The learning rate for training $G$ and $D$ are $10^{-3}$ and $10^{-4}$, respectively. We update both networks every 10 frames using 10 iterations. Our VITAL tracker runs on a PC with an i7 3.6GHz CPU and a Tesla K40c GPU with the MatConvNet toolbox \cite{Vedaldi-mm15-matconvnet} and the average speed is 1.5 FPS.

{\flushleft \bf Evaluation Metrics.} We follow the standard evaluation approaches. In the OTB-2013 and OTB-2015 datasets we use the one-pass evaluation (OPE) with precision and success plots metrics. The precision metric measures the frame locations rate within a certain threshold distance from ground truth locations. The threshold distance is set as 20 pixels. The success plot metric is set to measure the overlap ratio between the predicted bounding boxes and the ground truth. In the VOT-2016 dataset~\cite{kristan-eccvw16-vot}, we measure the performance in terms of Expected Average Overlap (EAO), Accuracy Ranks (Ar) and Robustness Ranks (Rr).

{\flushleft \bf Ablation Studies.}
In VITAL, we train the classifier using the diversified positive samples with a cost sensitive loss. To validate the effectiveness of each component, we first implement a baseline algorithm by not enabling the adversarial training and using the standard cross entropy loss. We implement three alternative approaches based on the baseline algorithm. First, we train the classifier by generating random masks. Second, we train the classifier using adversarial learning (i.e., GAN). Third, we train the classifier using adversarial learning with the cost sensitive loss. Fig. \ref{fig:ablation} shows the results on the OTB-2013 dataset. We observe that using random masks deteriorates the classifier and results in inferior performance. It is because the spatial discriminative and temporal robust features are blocked randomly, which degrades the classifier to focus on either. In contrast, the mask predicted by adversarial learning effectively exploits the most robust features by blocking partial discriminative features in individual frames. The cost sensitive loss further improves the performance. However, the improvement of the cost sensitive loss is not as salient as that of the adversarial learning.

\renewcommand{\tabcolsep}{0.1pt}
\begin{figure}[t]
\begin{center}
\begin{tabular}{cc}
\includegraphics[width=\swtwo]{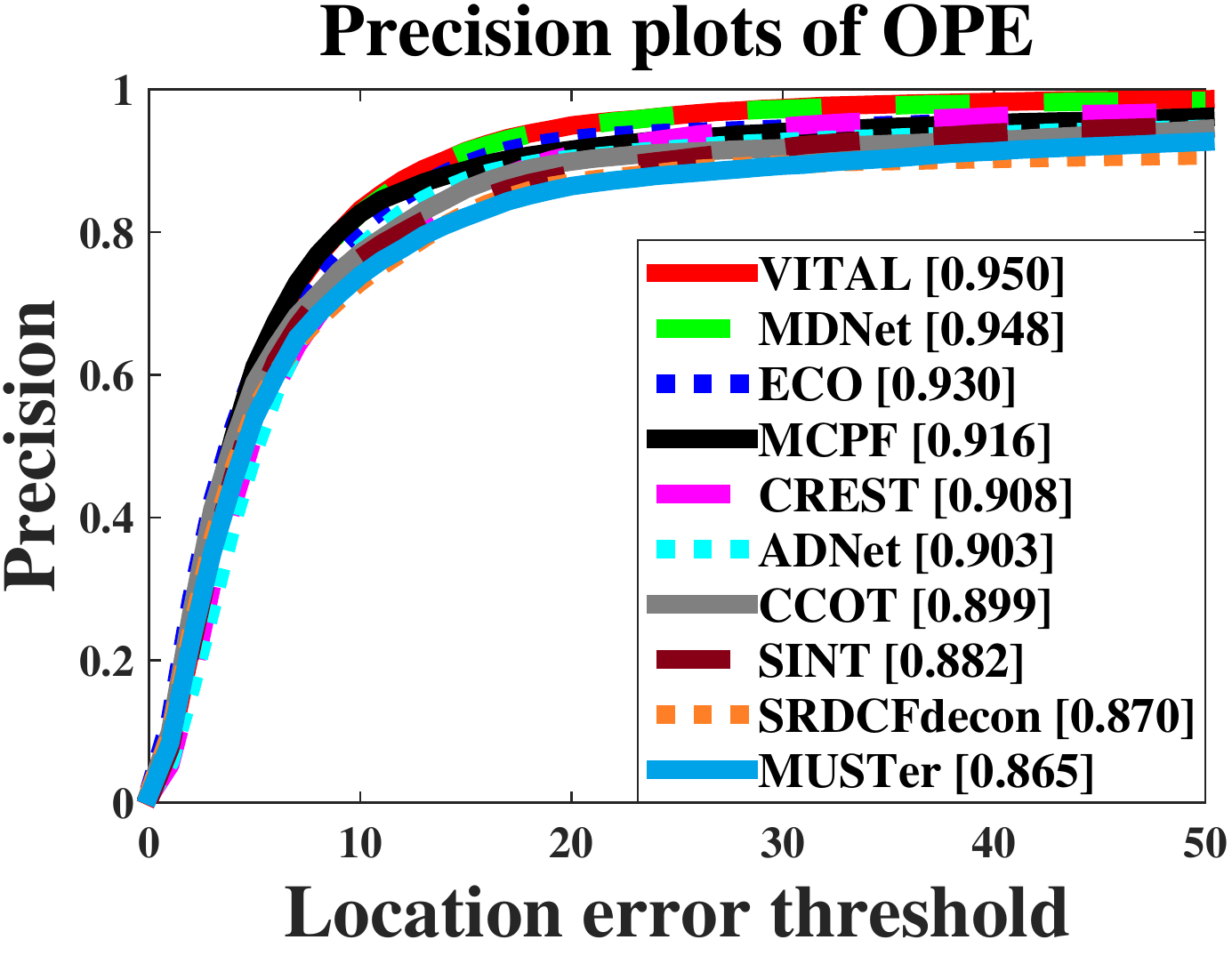}&
\includegraphics[width=\swtwo]{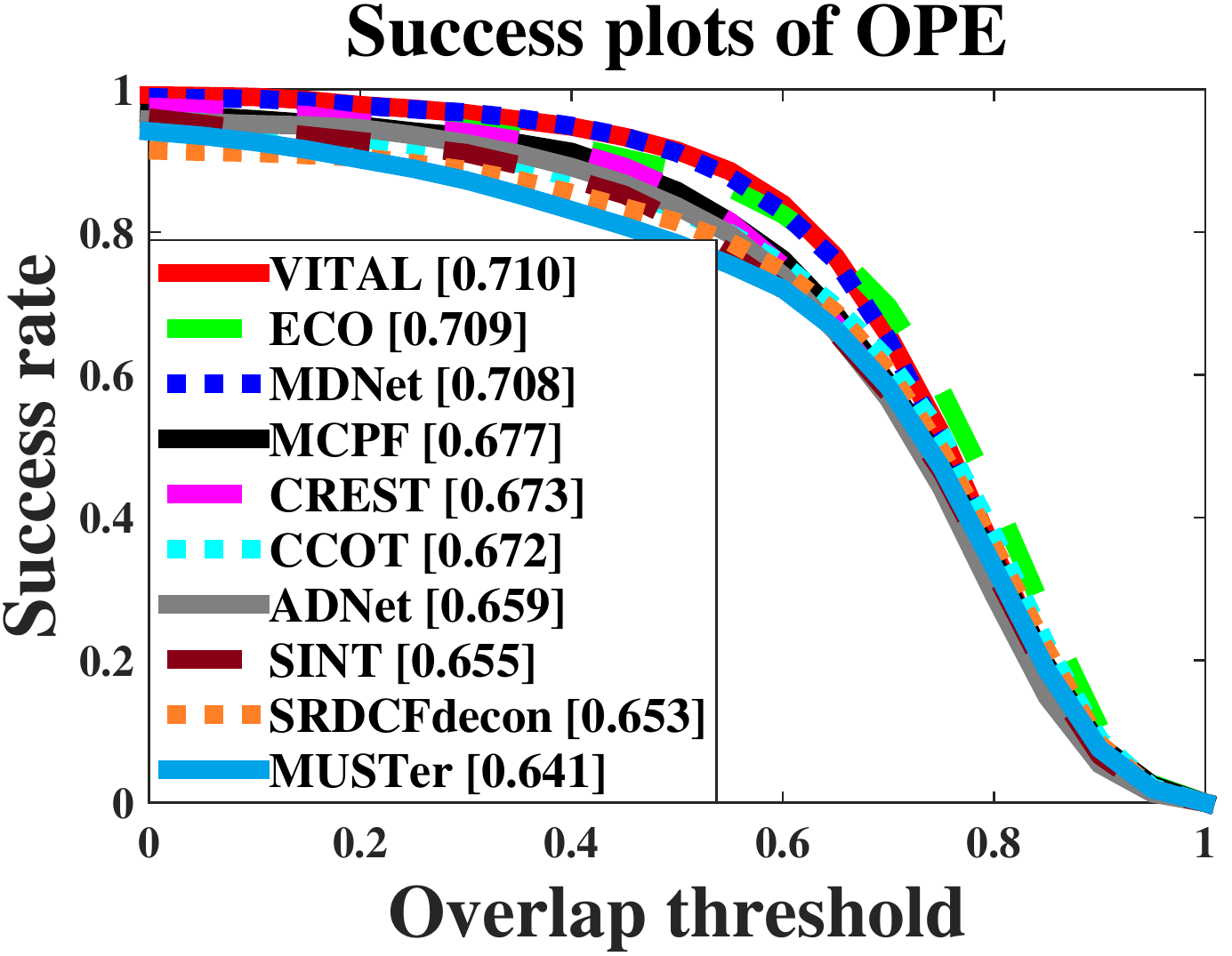}\\
\end{tabular}
\end{center}
\vspace{-5mm}
\caption{Precision and success plots on the OTB-2013 dataset using one-pass evaluation.}
\label{fig:otb2013}
\end{figure}

\renewcommand{\tabcolsep}{0.1pt}
\def\swfour{0.245\linewidth}
\begin{figure*}[t]
\begin{center}
\begin{tabular}{cccc}
\includegraphics[width=\swfour]{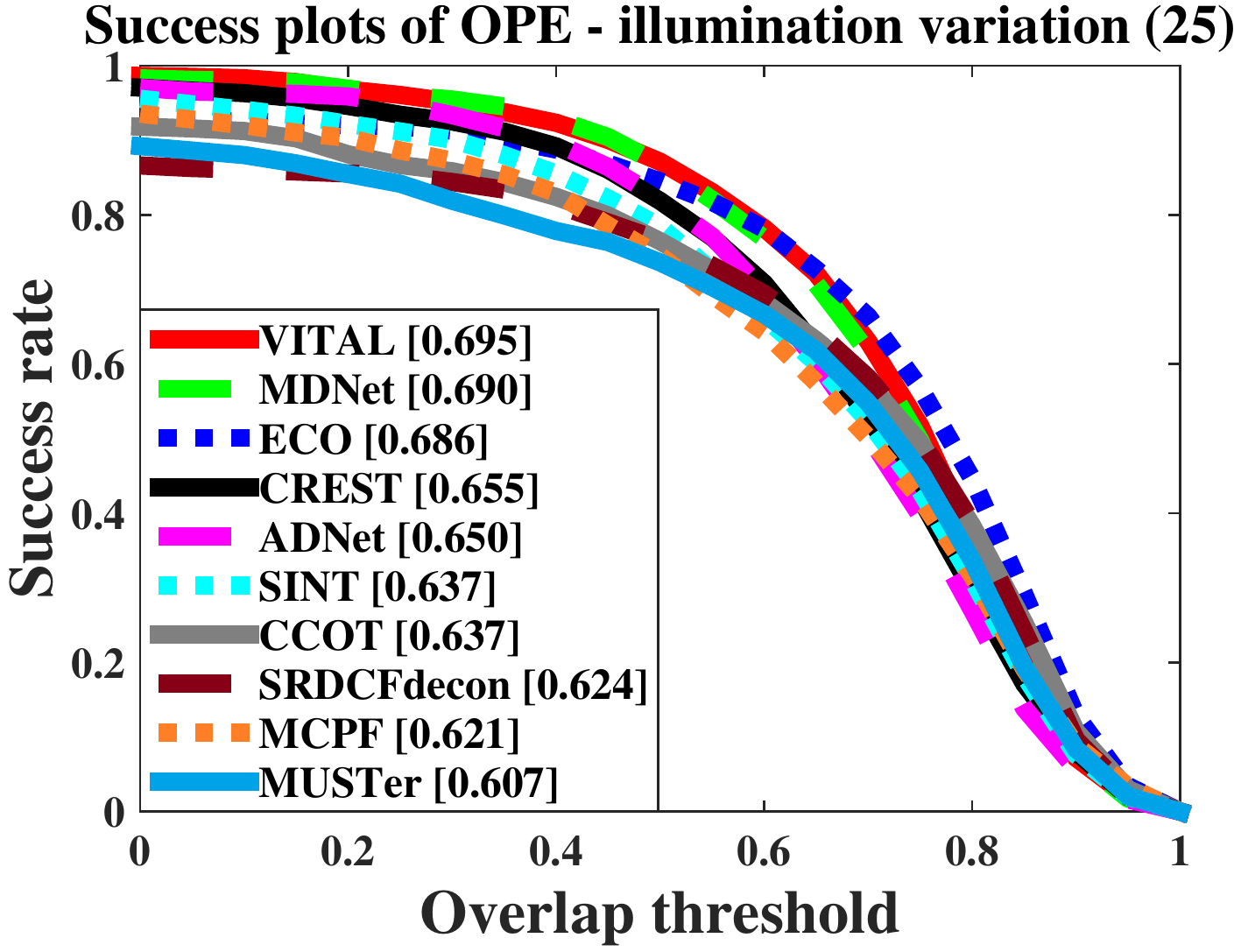}&
\includegraphics[width=\swfour]{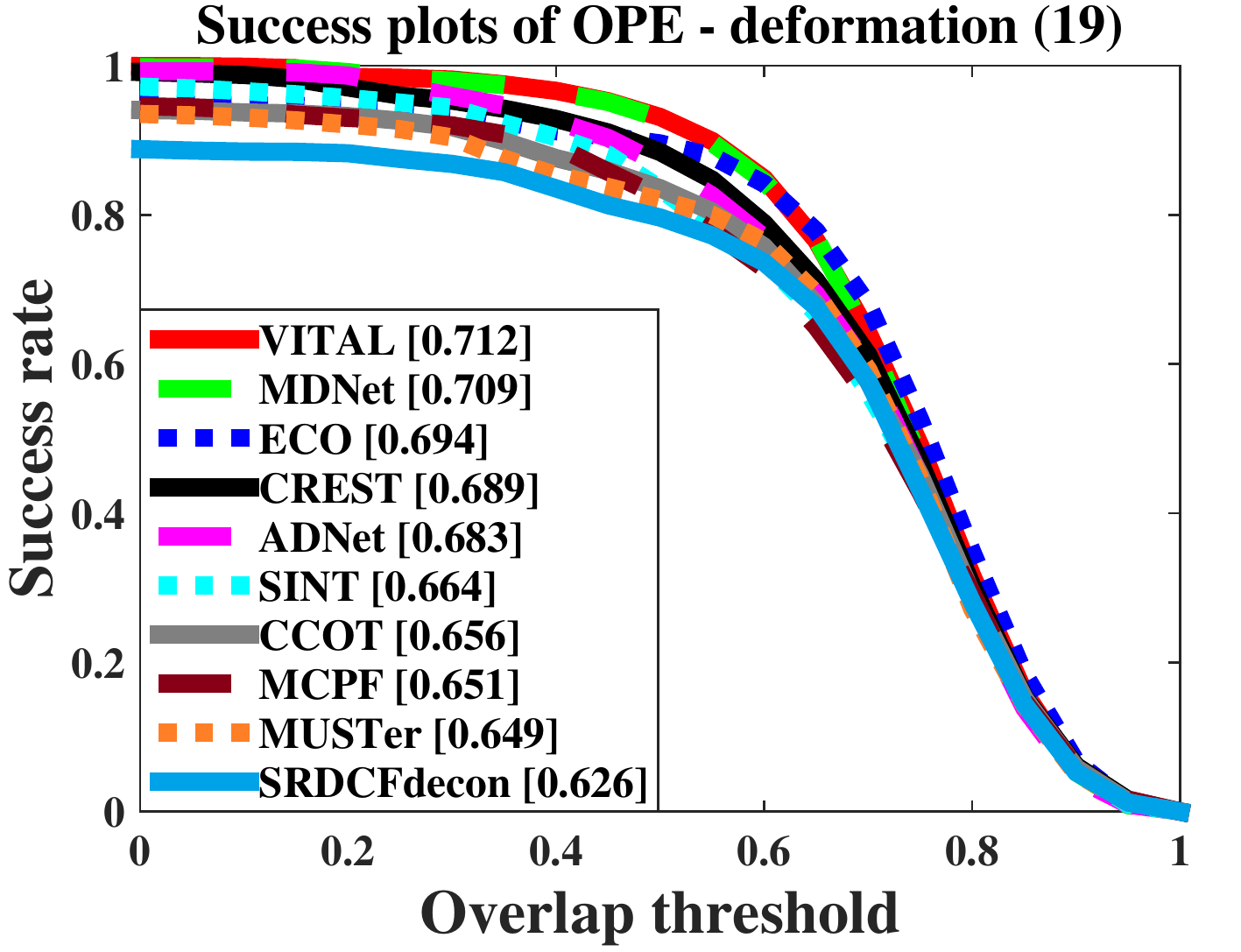}&
\includegraphics[width=\swfour]{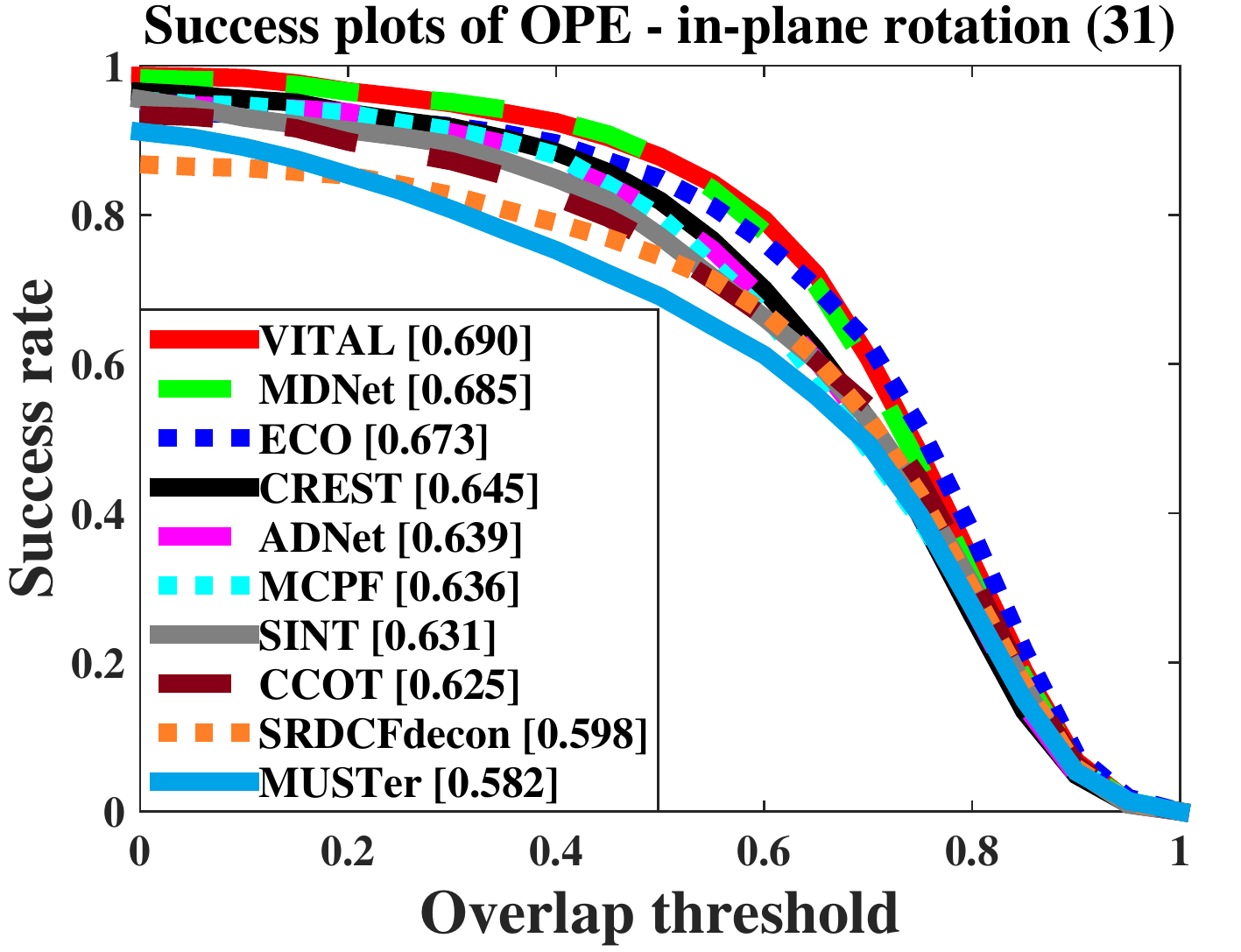}&
\includegraphics[width=\swfour]{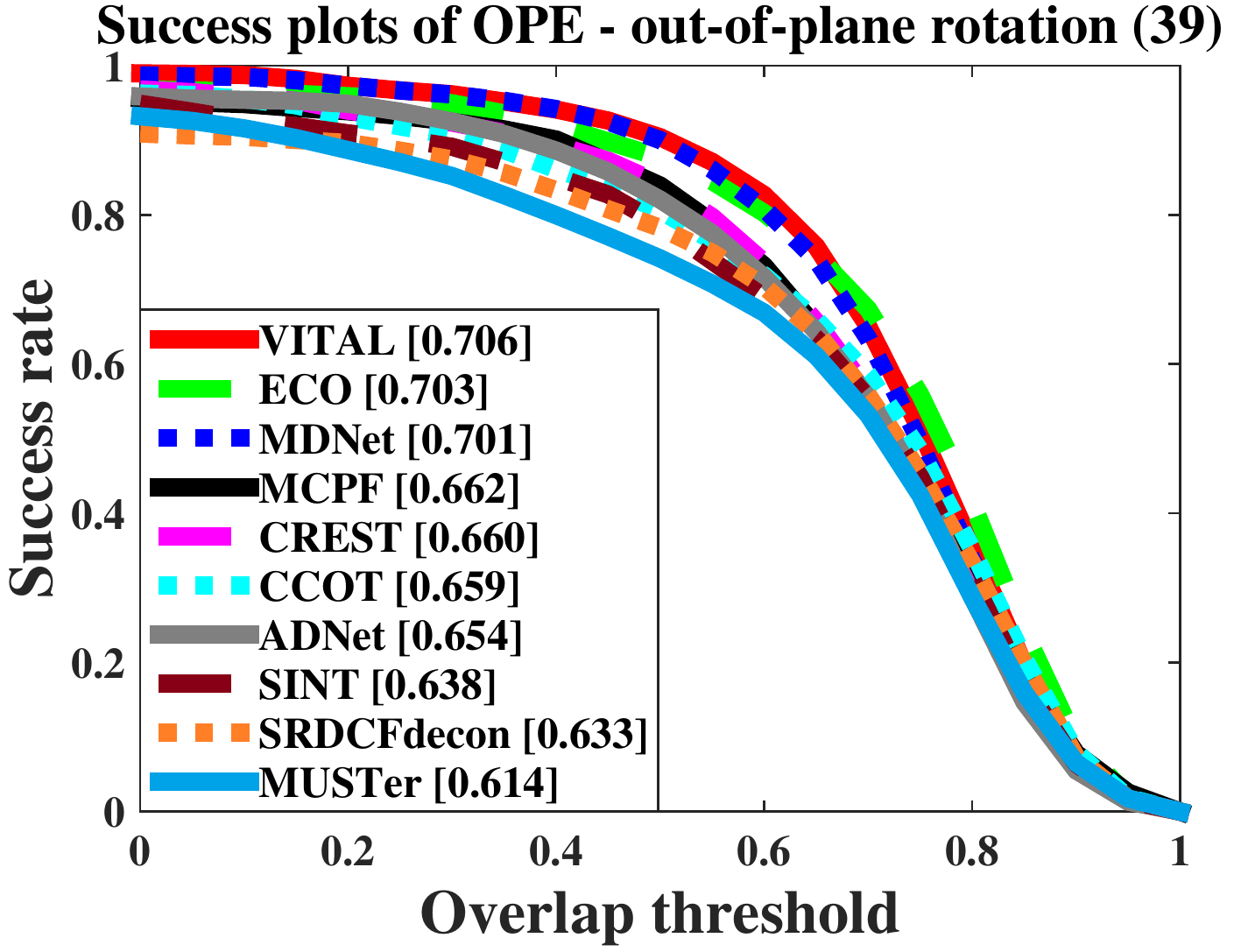}\\
\includegraphics[width=\swfour]{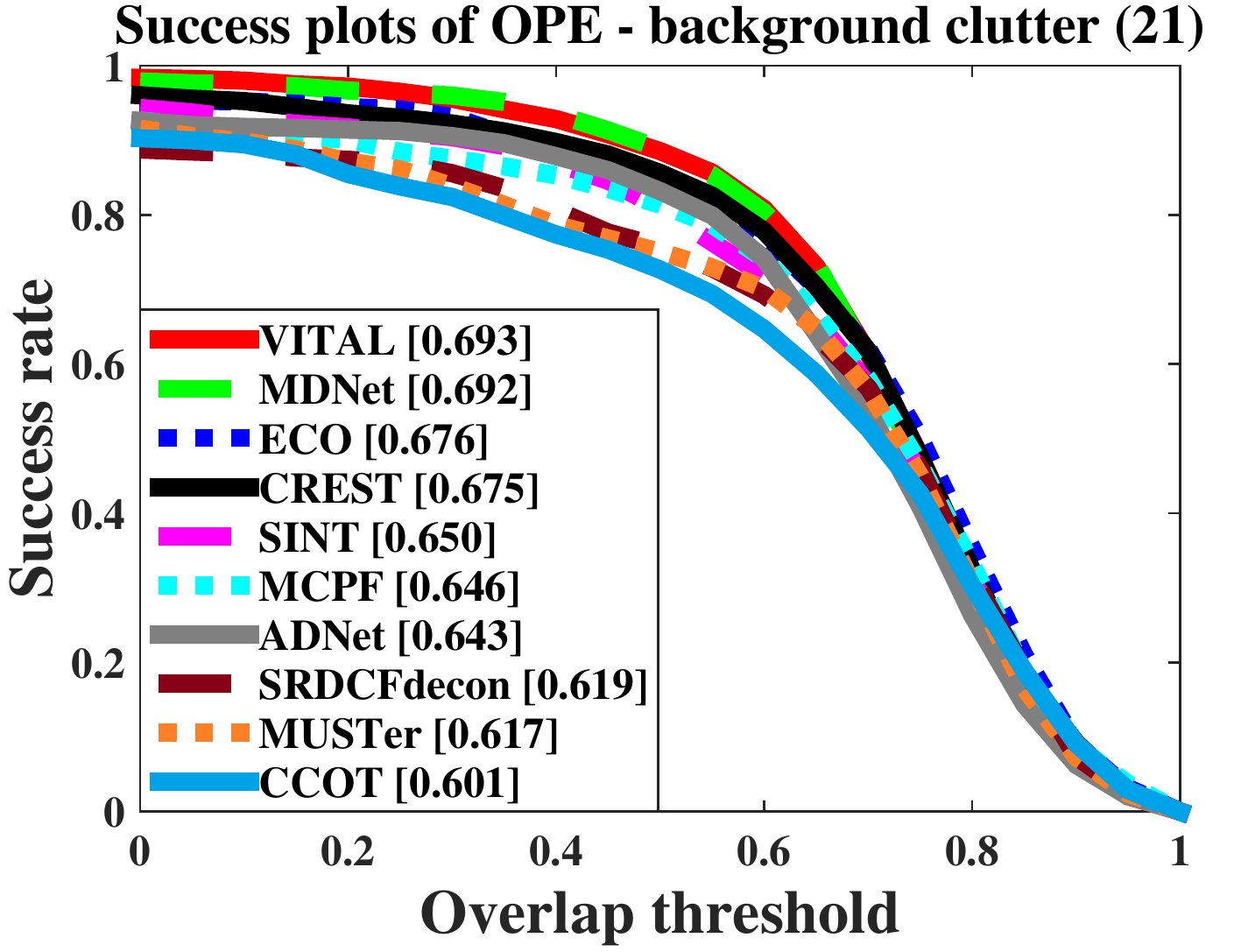}&
\includegraphics[width=\swfour]{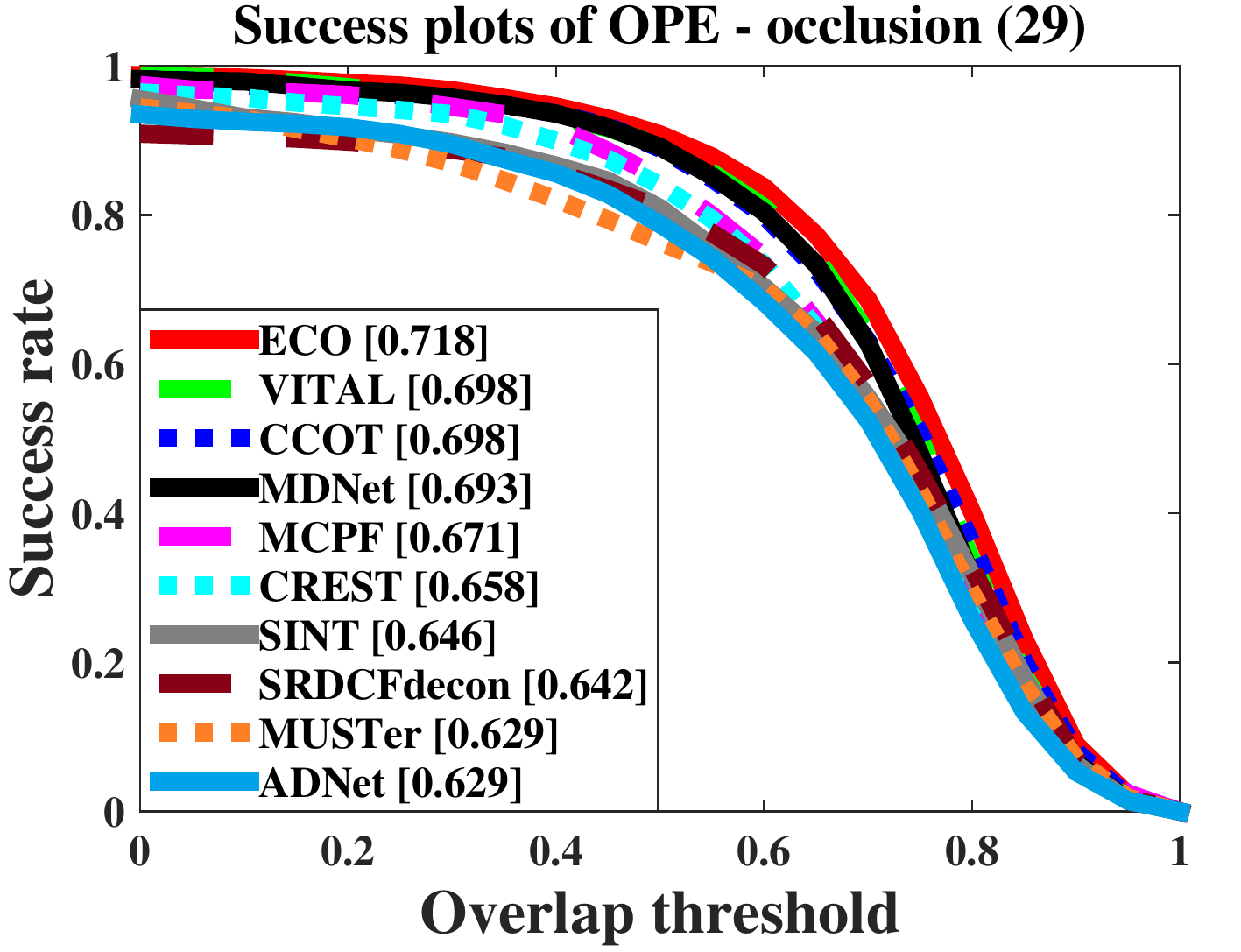}&
\includegraphics[width=\swfour]{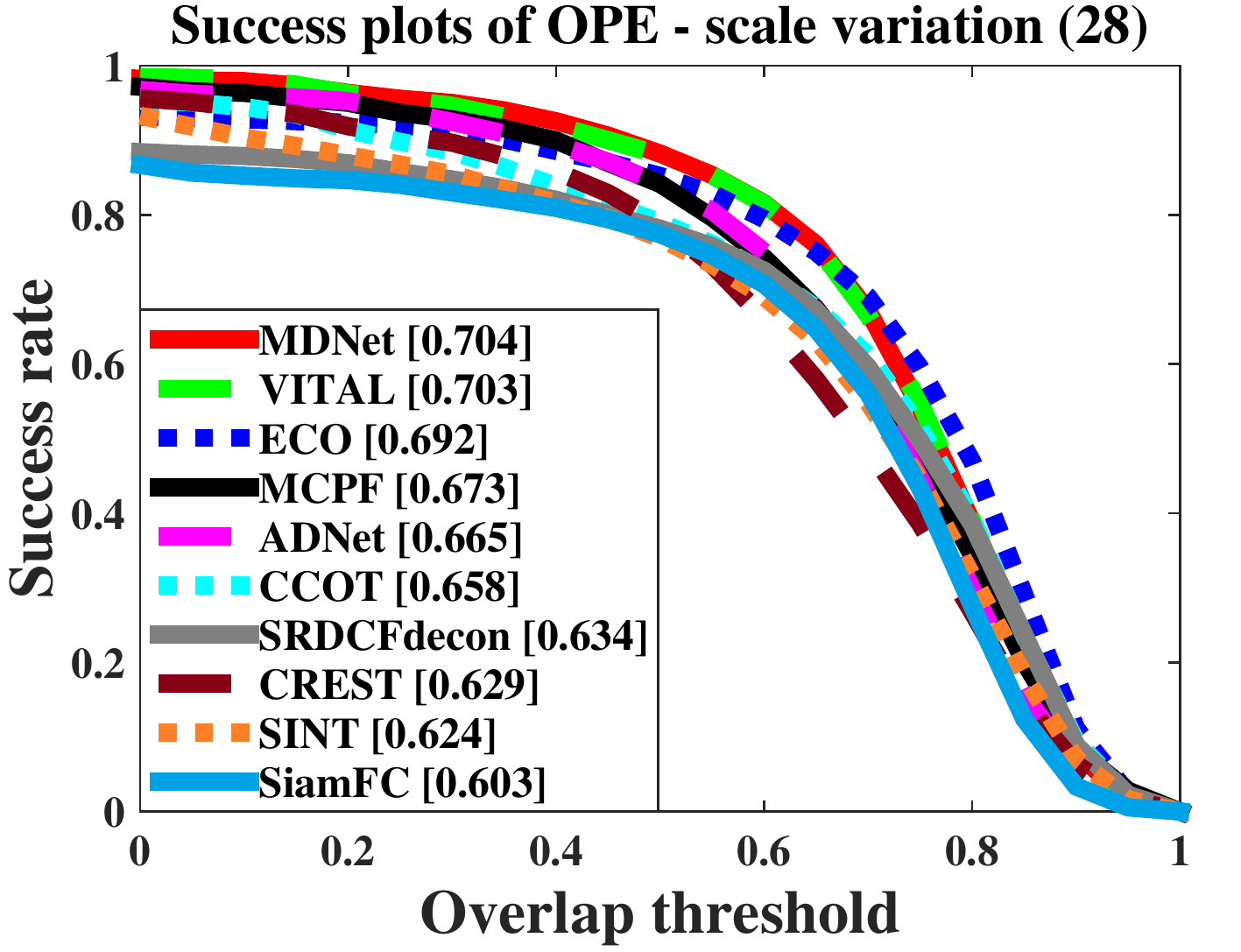}&
\includegraphics[width=\swfour]{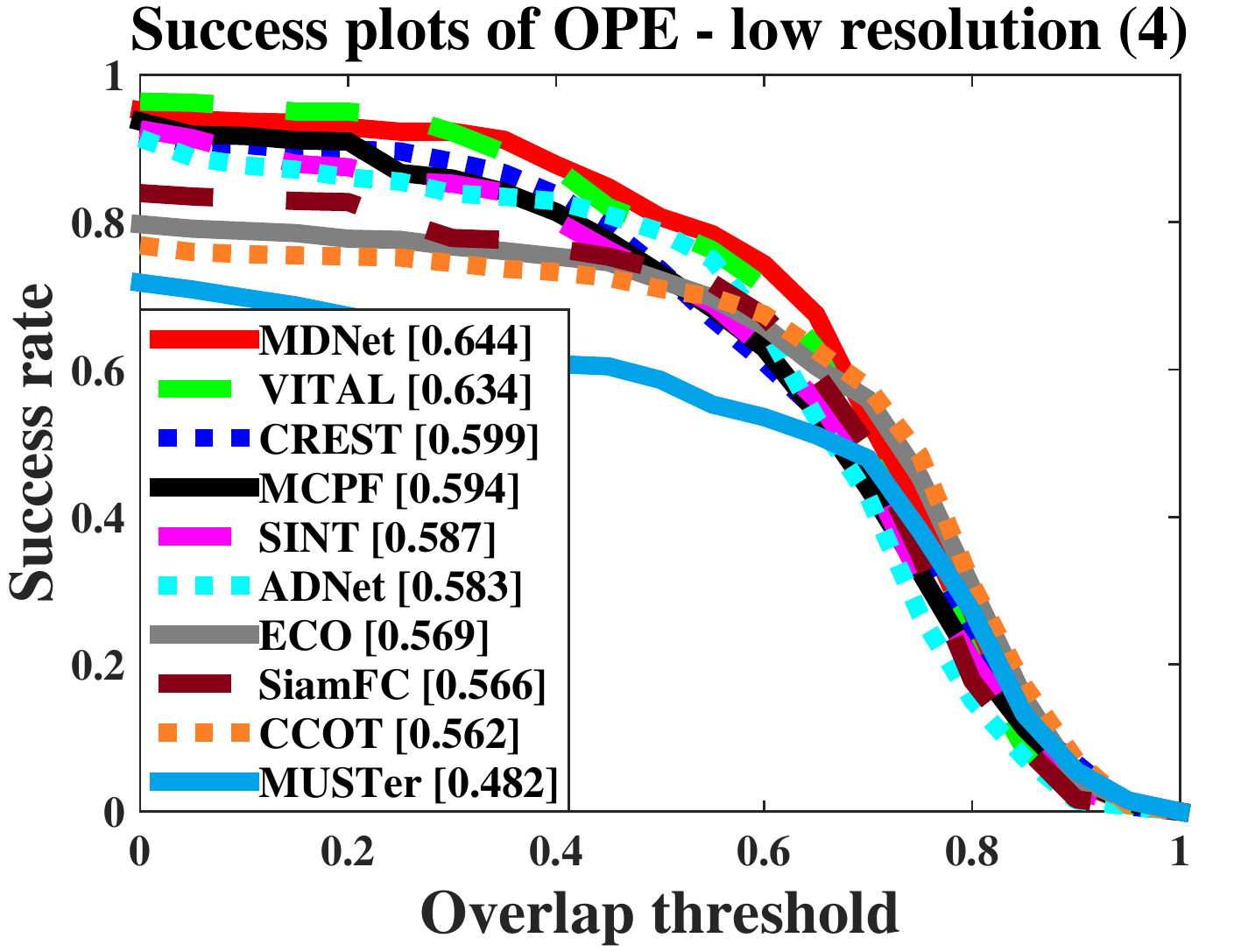}\\
\end{tabular}
\end{center}
\vspace{-5mm}
\caption{Overlap success plots over eight tracking challenges of illumination variation, deformation, in-plane rotation, out-of-plane rotation, background clutter, occlusion, scale variation and low resolution.}
\label{fig:otb2013attr}
\end{figure*}

\renewcommand{\tabcolsep}{0.1pt}
\begin{figure}[t]
\begin{center}
\begin{tabular}{cc}
\includegraphics[width=\swtwo]{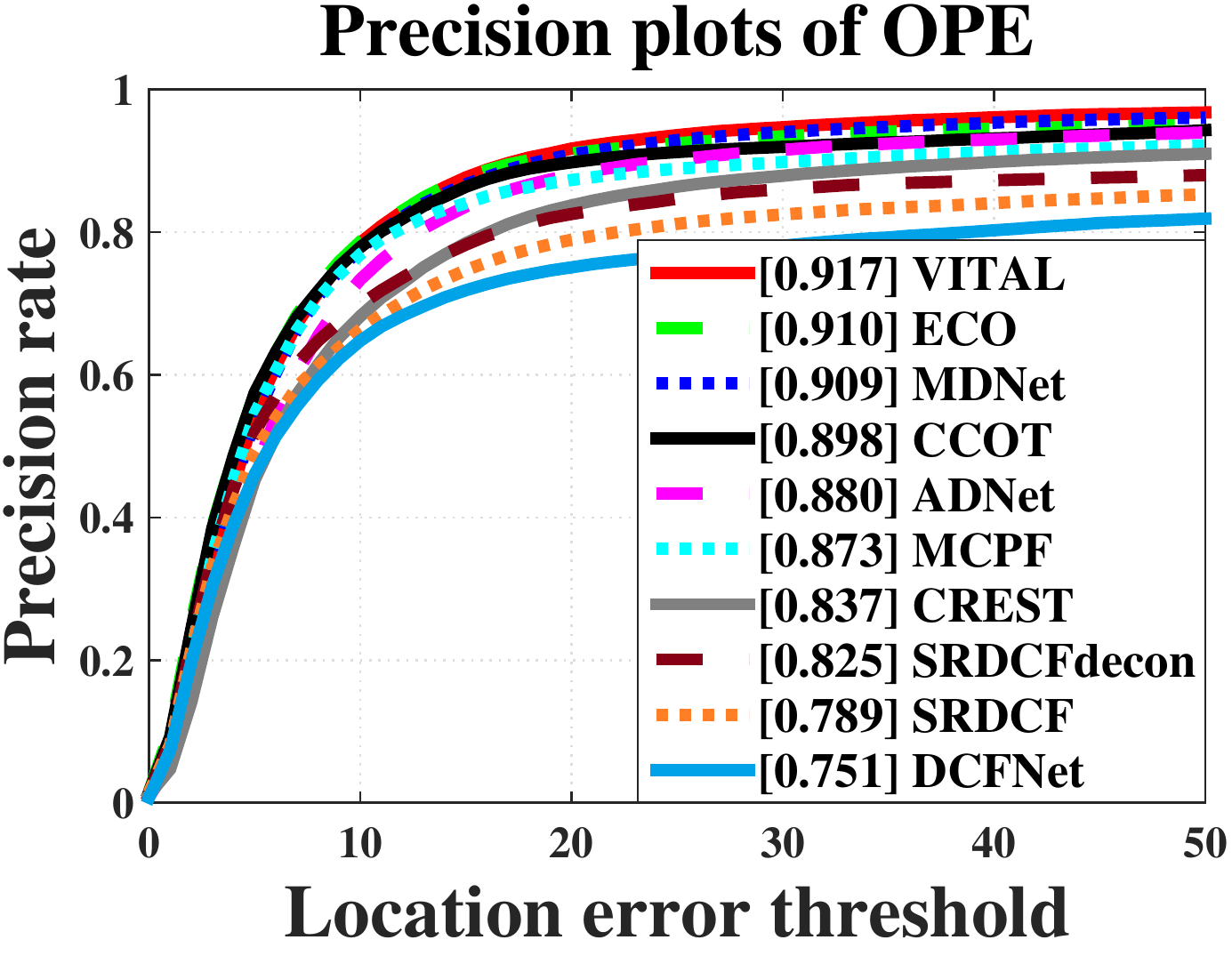}&
\includegraphics[width=\swtwo]{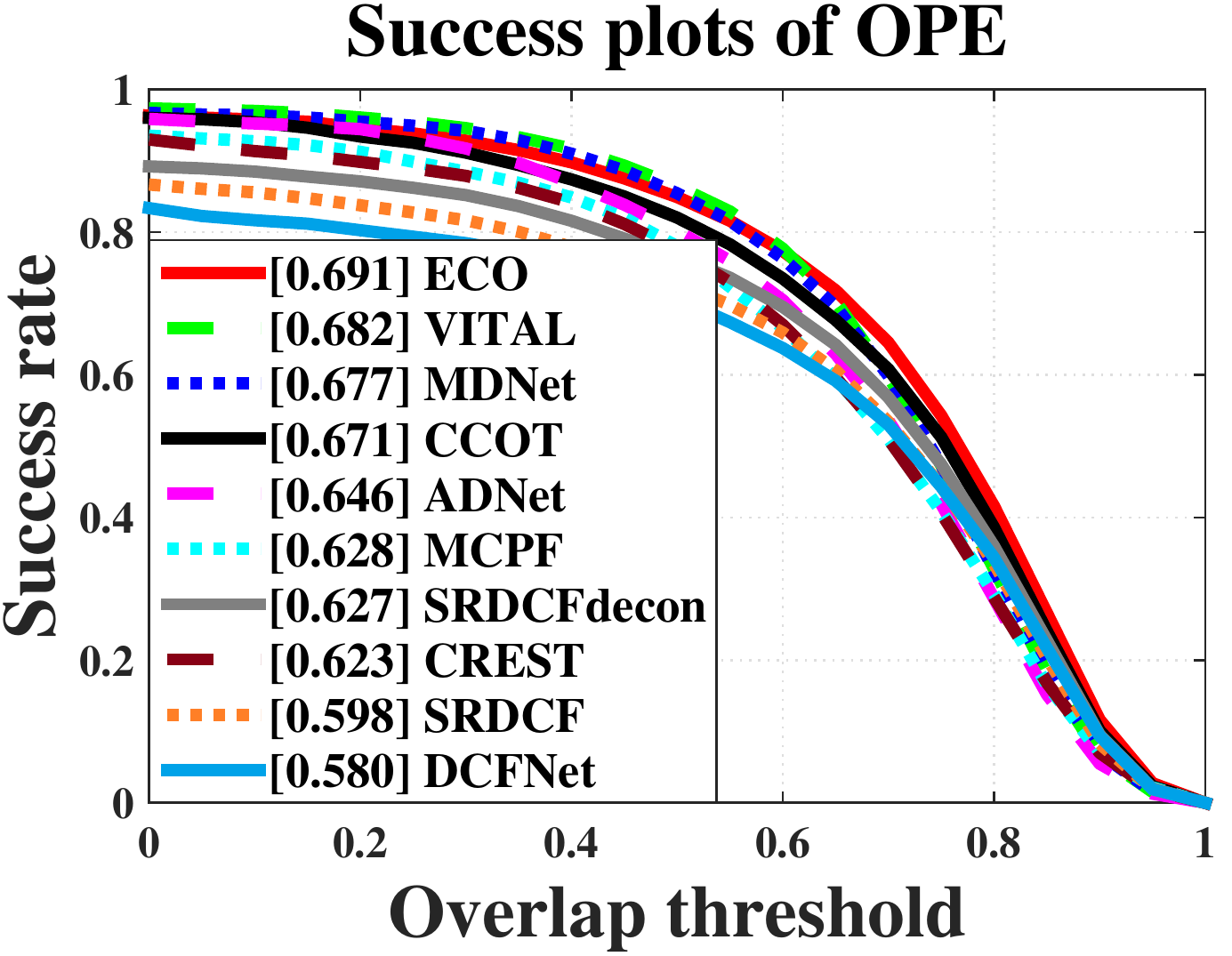}\\
\end{tabular}
\end{center}
\vspace{-5mm}
\caption{Precision and success plots on the OTB-2015 dataset using one-pass evaluation.}
\label{fig:otb2015}
\end{figure}

{\flushleft \bf OTB-2013 Dataset.}
We compare VITAL with 29 trackers from the OTB-2013 benchmark \cite{wu-cvpr13-otb} and other 28 state-of-the-art trackers including DSST \cite{martin-bmvc14-accurate}, KCF \cite{Henriques-eccv12-DCF}, TGPR \cite{gao-eccv14-transfer}, MEEM \cite{zhang-eccv14-meem}, RPT \cite{li-cvpr15-reliable}, LCT \cite{ma-cvpr15-lct}, MUSTer \cite{hong-cvpr15-muster}, HCFT \cite{chao-iccv15-HCF}, FCNT \cite{wang-iccv15-visual}, SRDCF \cite{martin-iccv15-learning}, CNN-SVM \cite{hong-icml15-cnnsvm}, DeepSRDCF \cite{Danelljan-iccvw15-DeepSRDCF}, DAT \cite{possegger-cvpr15-defense}, Staple \cite{bertinetto-cvpr16-staple}, SRDCFdecon \cite{danelljan-CVPR16-adaptive}, CCOT \cite{martin-eccv16-beyond}, GOTURN \cite{held-eccv16-learning}, SINT \cite{tao-cvpr16-siamese}, SiamFC \cite{bertinetto-eccv16-fully}, HDT \cite{qi-cvpr16-hdt}, SCT \cite{choi-cvpr16-visual}, MDNet~\cite{nam-cvpr16-mdnet}, DLS-SVM~\cite{ning-cvpr16-object}, ADNet~\cite{yun-cvpr17-action}, ECO~\cite{martin-cvpr17-eco}, MCPF~\cite{zhang-cvpr17-mcpf}, CFNet~\cite{luca-cvpr17-end} and CREST~\cite{song-iccv17-CREST}. We evaluate all the trackers on 50 video sequences using the one-pass evaluation with distance precision and overlap success metrics.

Figure \ref{fig:otb2013} shows the results from all compared trackers. For presentation clarity, we only show the top 10 trackers. The numbers listed in the legend indicate the AUC overlap success and 20 pixel distance precision scores. Overall, our VITAL tracker performs favorably against state-of-art trackers in both distance precision and overlap success. Figure \ref{fig:otb2013attr} compares the performance under eight video attributes using one-pass evaluation. Our VITAL tracker handles large appearance variations well caused by deformation, in-plane and out-of-plane rotations. Compared to the representative tracking-by-detection tracker MDNet, we attribute our performance improvement by the diversified positive samples for training robust classifiers. The mask generated via adversarial learning captures a variety of object variations. It maskouts the discriminative features in individual frames while maintains the most robust features over a long temporal span. The advantage of exploiting the temporally robust features is clearly proved when dealing with occlusion. Through focusing on the persistently robust features, our VITAL tracker performs better than MDNet in a large margin. Meanwhile, our cost sensitive loss effectively decreases the loss from easy negative samples and forces the classifier to focus on hard ones. This facilitates discriminative classifiers to separate the target object from background. Our VITAL achieves leading performance in the presence of illumination variation and background clutter. However, for the low resolution sequences, our tracker does not perform as well as MDNet. This is because the target size of these sequences is small and the resolution of the weight masks predicted by adversarial learning is far low. For the scale variance sequence, the fixed size of weight mask cannot precisely maskout the discriminative features as the object size increases. Our future work will consider adaptively changing the size of the weight mask.

{\flushleft \bf OTB-2015 Dataset.}
We compare our VITAL tracker on the OTB-2015 benchmark \cite{wu-pami15-otb} with the state-of-the-art trackers.
Figure \ref{fig:otb2015} shows that our VITAL tracker overall performs well. The ECO tracker achieves the best result in overlap success, while our VITAL ranks first in distance precision. Since the OTB-2015 dataset contains more videos with large scale changes and low resolution, our VITAL tracker does not perform as well as ECO in overlap success.

\def\pp{\hspace{0mm}}
\renewcommand{\tabcolsep}{6pt}
\begin{table}[t]
\caption{Comparison with the state-of-the-art trackers on the VOT 2016 dataset. The results are presented in terms of expected average overlap (EAO), accuracy rank (Ar) and robustness rank (Rr).}
\vspace{-3mm}
\centering
       \begin{tabular}{cccccc}
        \toprule
        &ECO&CCOT&Staple&MDNet&VITAL\\
        \midrule
        EAO&0.374&0.331&0.295&0.257&0.323\\
        Ar&1.55&1.63&1.65&1.63&1.63\\
        Rr&1.57&1.70&2.67&2.4&2.17\\
        \bottomrule
       \end{tabular}
\label{tab:vot}
\vspace{-5mm}
\end{table}

\def\swthree{0.32\linewidth}
\def\swone{0.7\linewidth}
\begin{figure*}[t]
\begin{center}
\begin{tabular}{cc}
\begin{minipage}[t]{0.49\linewidth}
    \centering
    \renewcommand{\tabcolsep}{.1pt}
    \begin{tabular}{ccc}
    \vspace{-0.5mm}\includegraphics[width=\swthree]{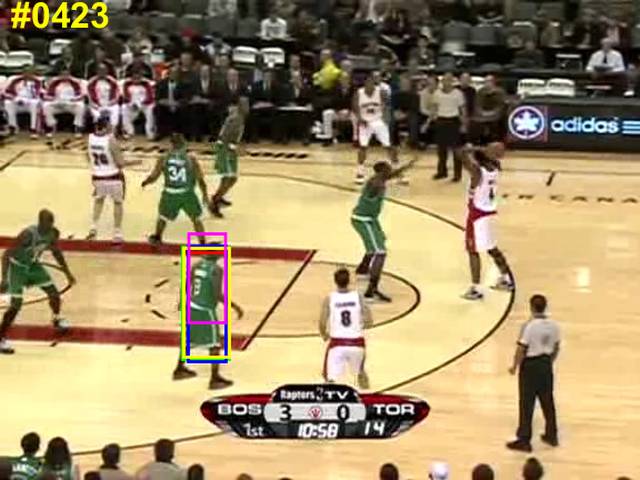}&
    \includegraphics[width=\swthree]{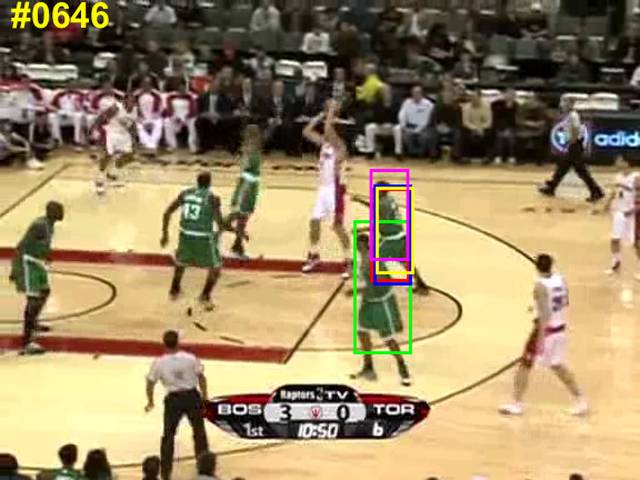}&
    \includegraphics[width=\swthree]{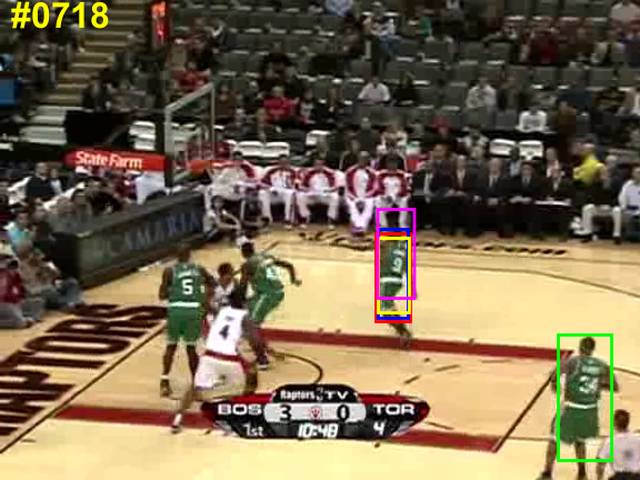}\\
    \vspace{-0.5mm}\includegraphics[width=\swthree]{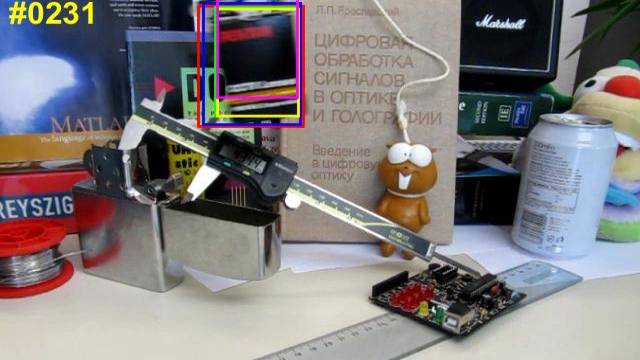}&
    \includegraphics[width=\swthree]{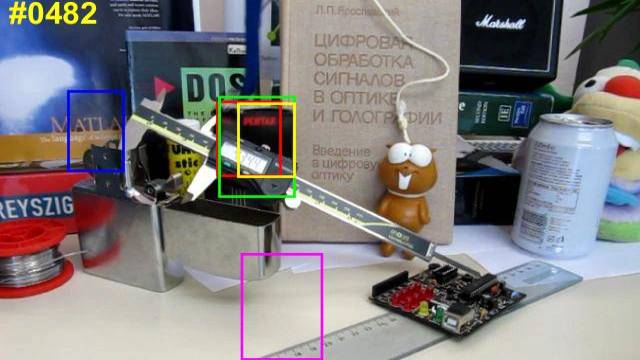}&
    \includegraphics[width=\swthree]{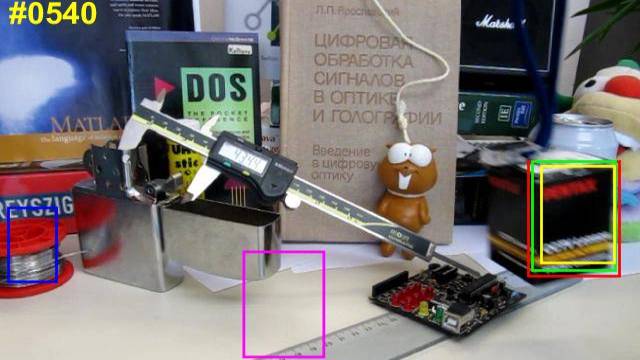}\\
    \vspace{-0.5mm}\includegraphics[width=\swthree]{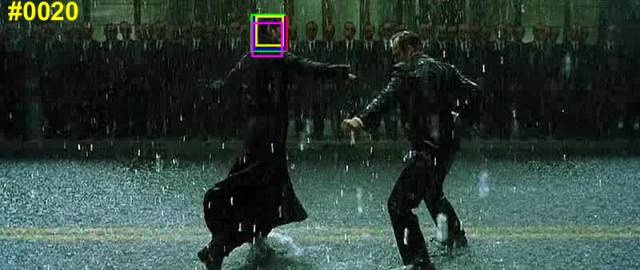}&
    \includegraphics[width=\swthree]{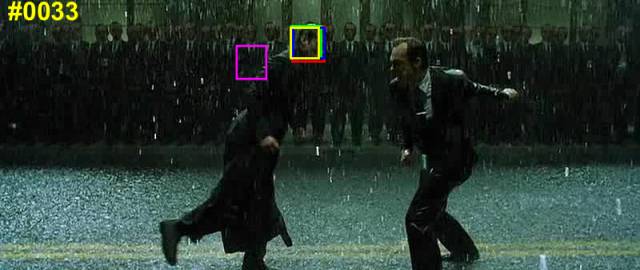}&
    \includegraphics[width=\swthree]{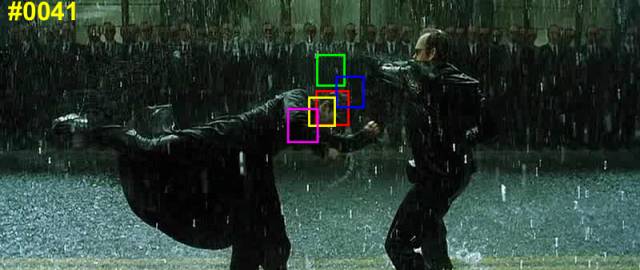}\\
    \vspace{-0.5mm}\includegraphics[width=\swthree]{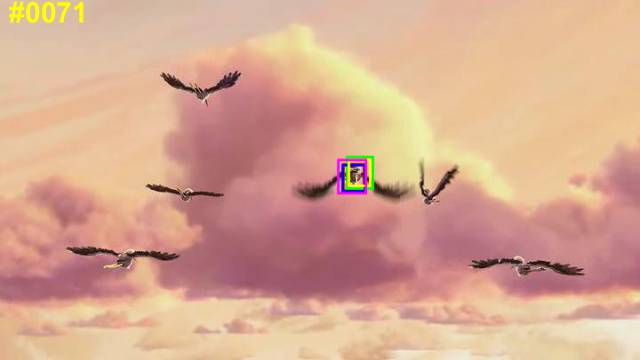}&
    \includegraphics[width=\swthree]{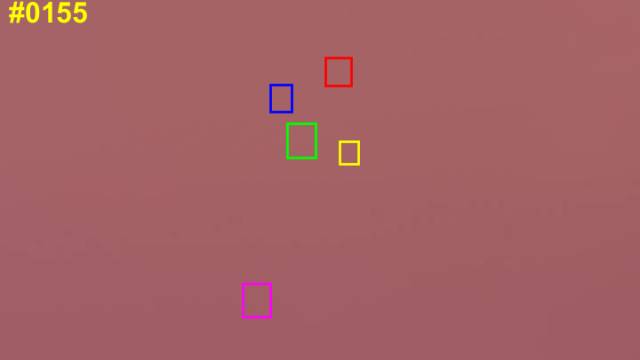}&
    \includegraphics[width=\swthree]{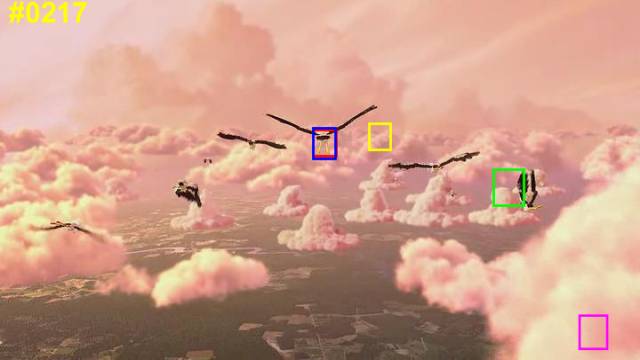}\\
    \vspace{-0.5mm}\includegraphics[width=\swthree]{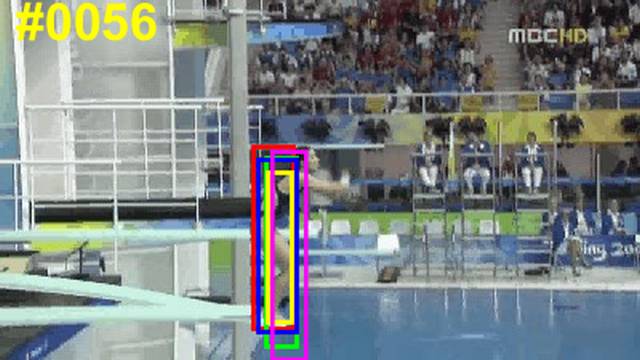}&
    \includegraphics[width=\swthree]{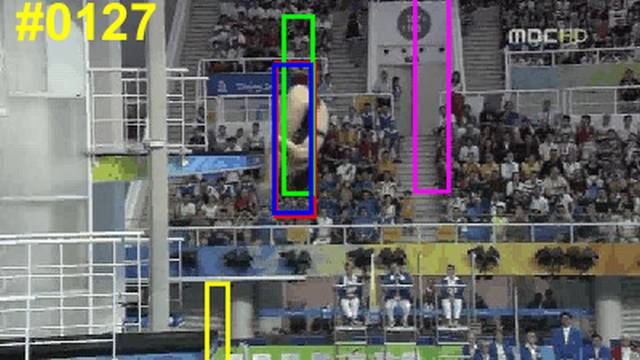}&
    \includegraphics[width=\swthree]{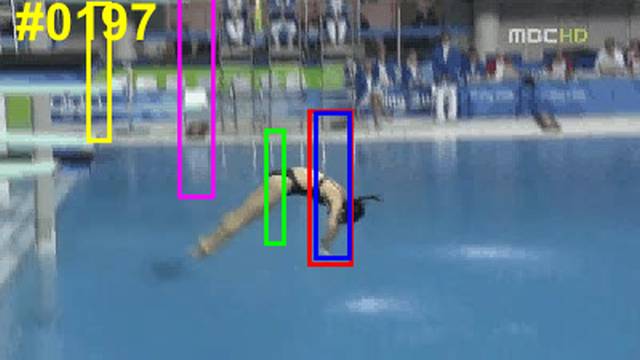}\\
    \includegraphics[width=\swthree]{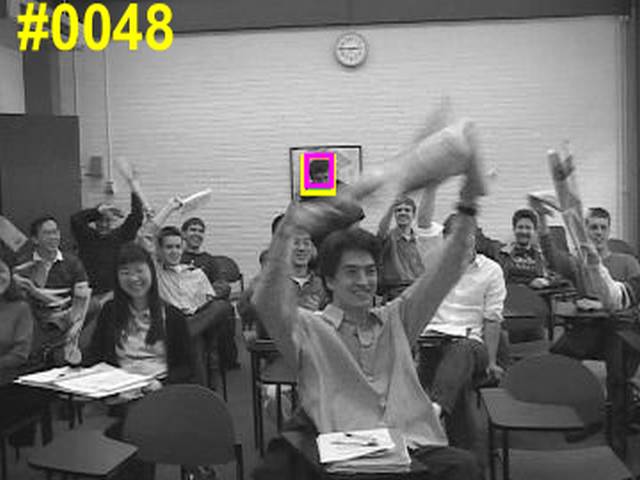}&
    \includegraphics[width=\swthree]{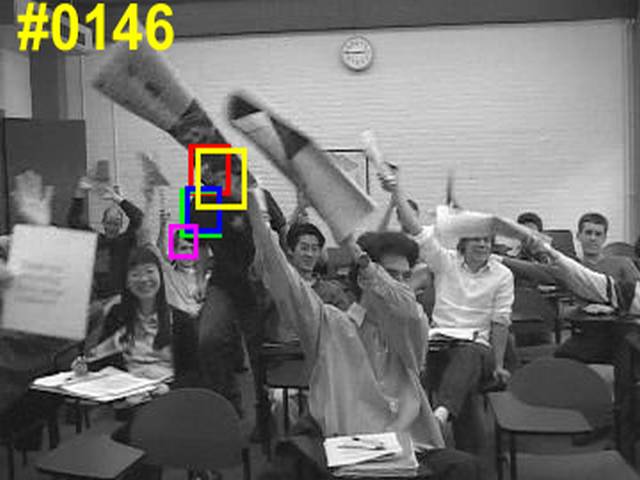}&
    \includegraphics[width=\swthree]{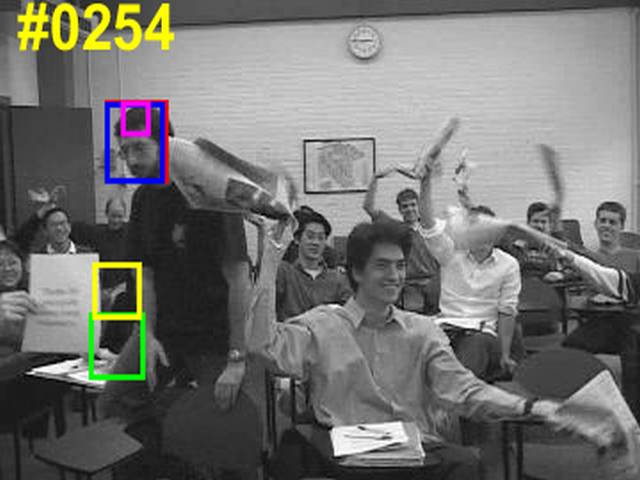}\\
    \end{tabular}
\end{minipage}
\begin{minipage}[t]{0.49\linewidth}
    \centering
    \renewcommand{\tabcolsep}{.1pt}
    \begin{tabular}{ccc}
    \vspace{-0.5mm}\includegraphics[width=\swthree]{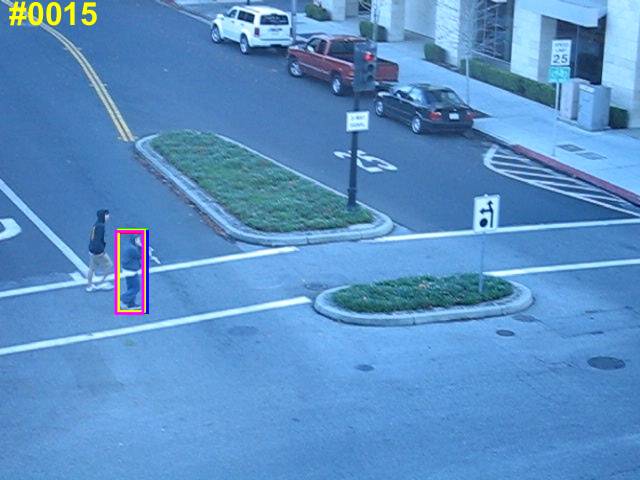}&
    \includegraphics[width=\swthree]{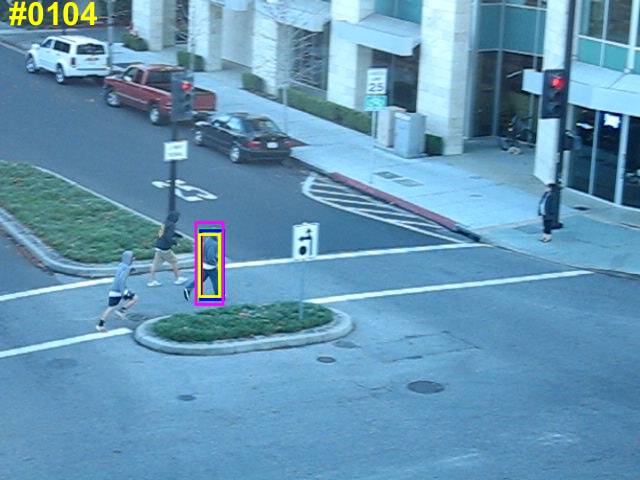}&
    \includegraphics[width=\swthree]{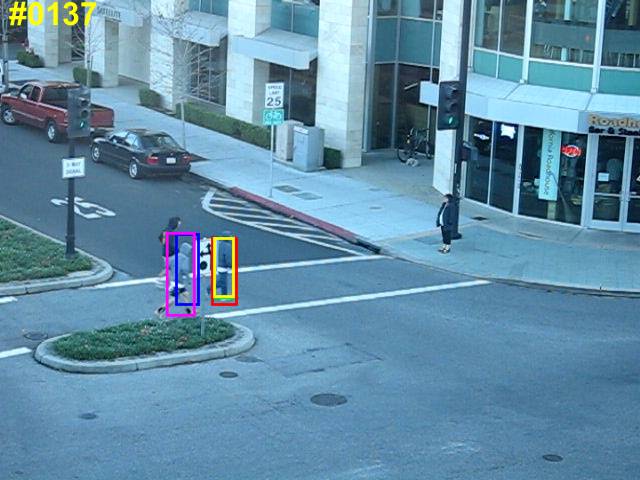}\\
    \vspace{-0.5mm}\includegraphics[width=\swthree]{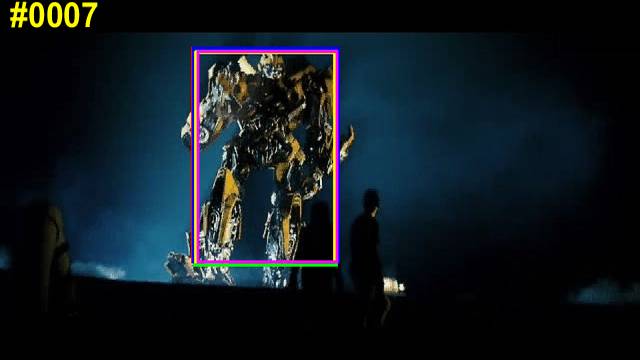}&
    \includegraphics[width=\swthree]{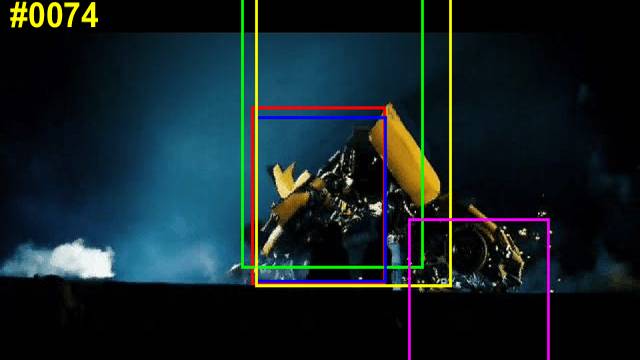}&
    \includegraphics[width=\swthree]{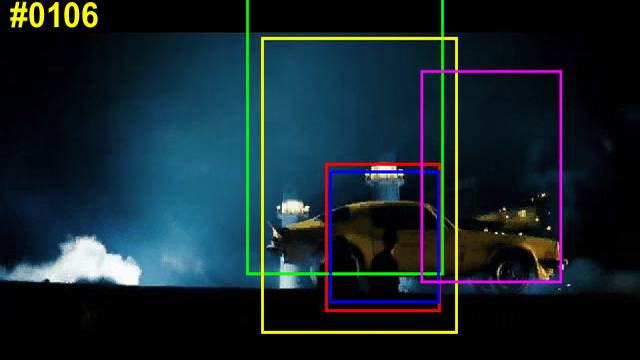}\\
    \vspace{-0.5mm}\includegraphics[width=\swthree]{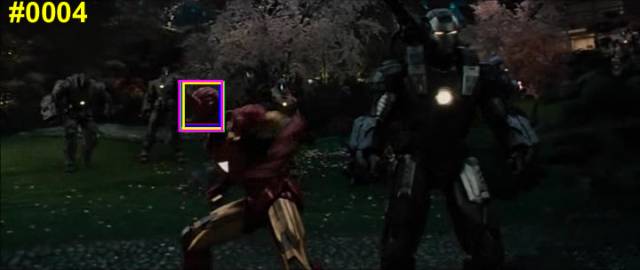}&
    \includegraphics[width=\swthree]{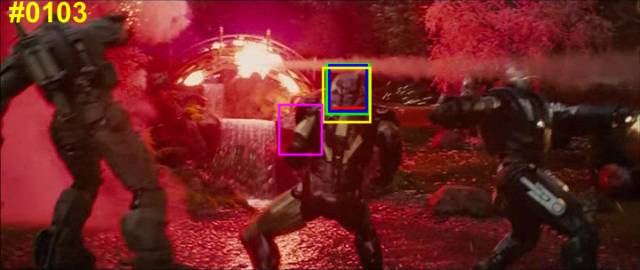}&
    \includegraphics[width=\swthree]{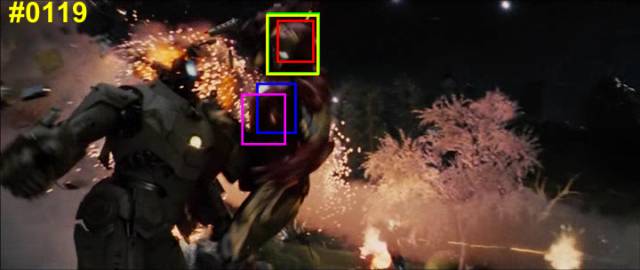}\\
    \vspace{-0.5mm}\includegraphics[width=\swthree]{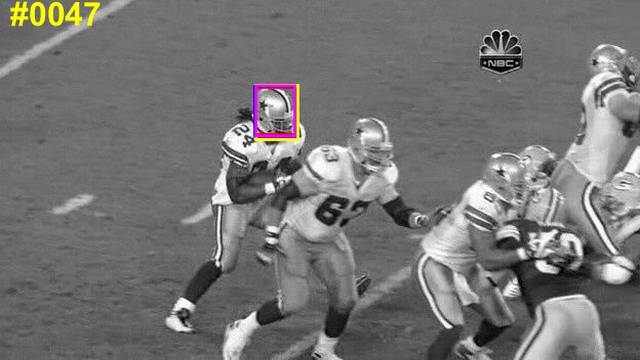}&
    \includegraphics[width=\swthree]{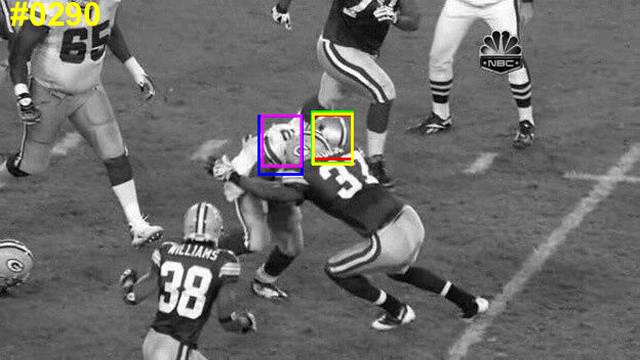}&
    \includegraphics[width=\swthree]{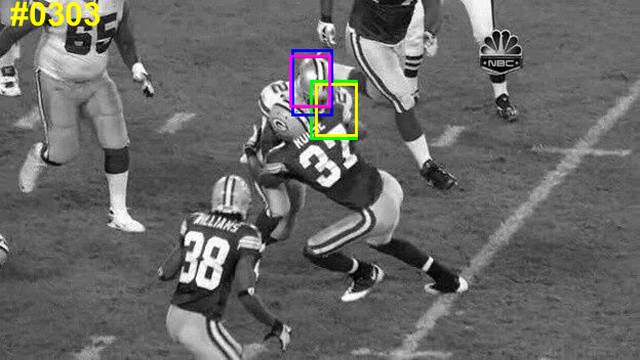}\\
    \vspace{-0.5mm}\includegraphics[width=\swthree]{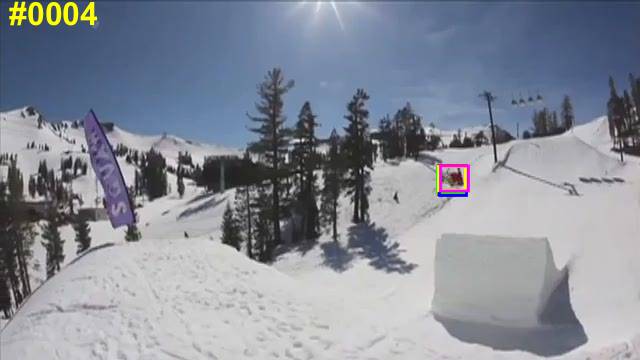}&
    \includegraphics[width=\swthree]{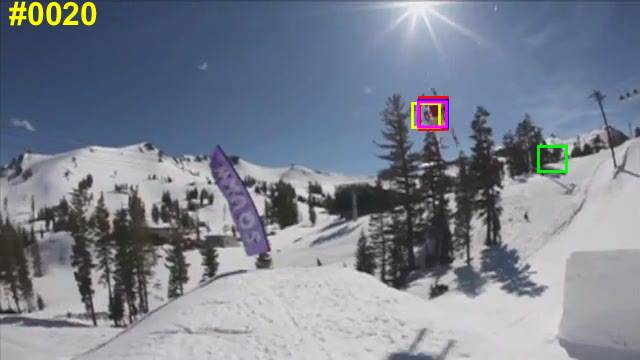}&
    \includegraphics[width=\swthree]{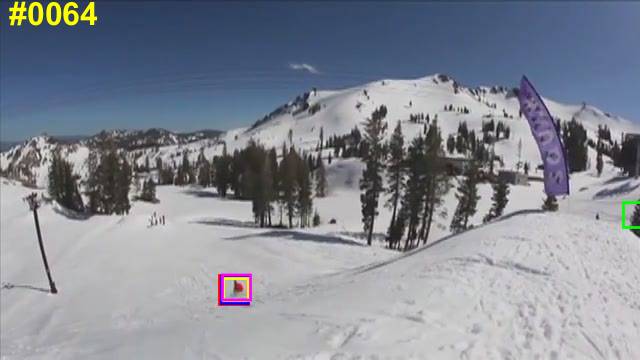}\\
    \includegraphics[width=\swthree]{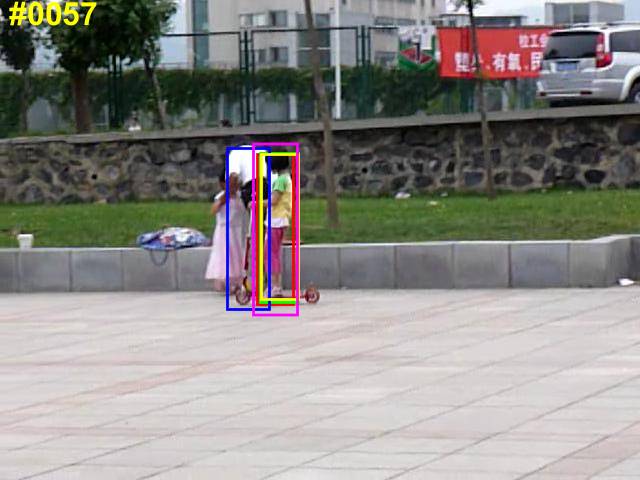}&
    \includegraphics[width=\swthree]{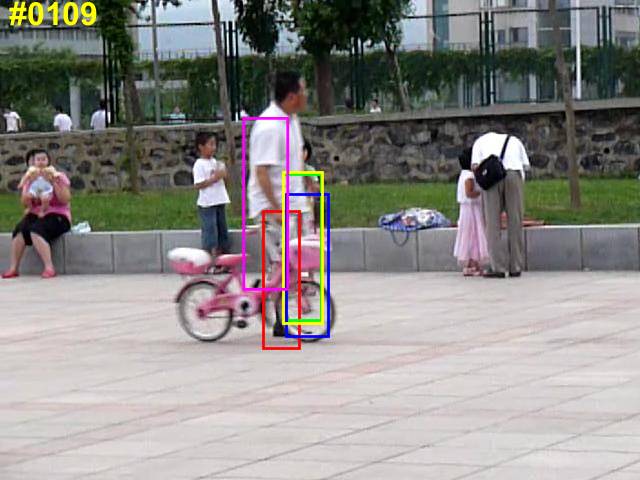}&
    \includegraphics[width=\swthree]{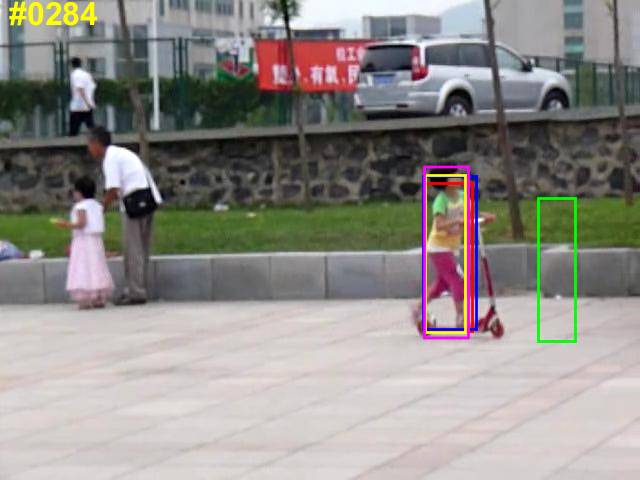}\\
    \end{tabular}
\end{minipage}\\
\end{tabular}
\begin{tabular}{c}
\includegraphics[width=\swone]{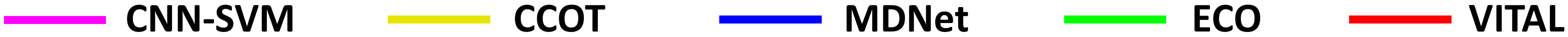}\\
\end{tabular}
\end{center}
\vspace{-5mm}
\caption{Qualitative evaluation of our VITAL tracker, CNN-SVM \cite{hong-icml15-cnnsvm}, CCOT \cite{martin-eccv16-beyond}, MDNet \cite{nam-cvpr16-mdnet}, ECO \cite{martin-cvpr17-eco} on 12 challenging sequences (from left to right and top to down: \emph{Basketball}, \emph{Human4}, \emph{Box}, \emph{Trans}, \emph{Matrix}, \emph{Ironman}, \emph{Bird1}, \emph{Football}, \emph{Diving}, \emph{Skiing}, \emph{Freeman4} and \emph{Girl2}, respectively). Our VITAL tracker performs favorably against state-of-the-art.}
\label{fig:visual}
\end{figure*}

{\flushleft \bf VOT-2016 Dataset.} We compare our VITAL tracker with state-of-the-art trackers on the VOT-2016 benchmark, including Staple~\cite{bertinetto-cvpr16-staple}, MDNet~\cite{nam-cvpr16-mdnet}, CCOT~\cite{martin-eccv16-beyond} and ECO~\cite{martin-cvpr17-eco}. VOT-2016 report~\cite{kristan-eccvw16-vot} shows that the strict state-of-the-art bound is 0.251 under EAO metric. Trackers whose EAO value exceeds this bound is defined as state-of-the-art. Table \ref{tab:vot} shows that ECO performs best under the EAO metric. The performance of VITAL is comparable to that of CCOT and better than Staple and MDNet. According to the definition of the VOT report, all these trackers are state-of-the-art.

{\flushleft \bf Qualitative Evaluation.}
Fig. \ref{fig:visual} qualitatively compare the results of the top performing trackers: CNN-SVM \cite{hong-icml15-cnnsvm}, CCOT \cite{martin-eccv16-beyond}, MDNet \cite{nam-cvpr16-mdnet}, ECO \cite{martin-cvpr17-eco} and VITAL on 12 challenging sequences. In a majority of these sequences, CNN-SVM fails to locate the target objects or estimates scale incorrectly because of the limited performance of the SVM classifier. MDNet improves CNN-SVM through an end-to-end CNN network formulation. It performs well on deformation (\emph{Trans}), low resolution (\emph{Skiing}) and fast motion (\emph{Diving}). However, the classifier of MDNet is trained to focus on the discriminative features from individual frames, which may lead to overfitting in the presence of noisy update. It does not perform well in handling out-of-plane rotation (\emph{Ironman}) and occlusion (\emph{Human4}). The correlation filter based trackers (i.e., CCOT and ECO) extract CNN features and learn correlation filters independently. They do not take full advantage of the end-to-end deep architecture.
In contrast, our VITAL tracker emphasizes on the most temporally robust features. The adversarial learning scheme makes the classifier aware a variety of appearance changes. The cost sensitive loss mines hard negative samples to further facilitate classifier learning. Our tracker VITAL performs favorably against state-of-the-art trackers.

\section{Conclusion}
In this paper we integrate adversarial learning into the tracking-by-detection framework to reduce overfitting on single frames. We adaptively dropout the discriminative features in single frame which draws the classifier attention. It enables the classifier to focus on the temporal robust features which are originally diminished during the training process. The adaptive dropout is achieved via adversarial learning to predict discriminative features according to different inputs. It enriches the target appearances in the feature space and augment the positive samples. Meanwhile, we use the cost sensitive loss to reduce the effect from easy negative samples.
Extensive experiments on benchmarks demonstrate that our VITAL tracker performs favorably against state-of-the-art trackers.

\newpage

{\small
\bibliographystyle{ieee}
\bibliography{ref}
}

\end{document}